\documentclass{article}
\usepackage[table]{xcolor}
\usepackage[english]{babel}
\usepackage[natbibapa]{apacite}
\usepackage[letterpaper,top=2cm,bottom=2cm,left=3cm,right=3cm,marginparwidth=1.75cm]{geometry}
\usepackage[colorlinks=true, citecolor=gray!160]{hyperref}
\usepackage{amsmath}
\usepackage{graphicx}
\usepackage{algorithm}
\usepackage{algpseudocode}
\usepackage{pgfplots}
\pgfplotsset{width=10cm,compat=1.9}
\usepackage{booktabs} % For formal tables
\usepackage{setspace}
\usepackage{subcaption}

\title{Enhancing Optimization Through Innovation: The Multi-Strategy Improved Black Widow Optimization Algorithm (MSBWOA)}
\author{Xin Xu}
\date{}

\begin{document}
\maketitle
\onehalfspacing

\begin{abstract}
This paper introduces a Multi-Strategy Improved Black Widow Optimization Algorithm (MSBWOA), designed to enhance the performance of the standard Black Widow Algorithm (BW) in solving complex optimization problems. The proposed algorithm integrates four key strategies: initializing the population using Tent chaotic mapping to enhance diversity and initial exploratory capability; implementing mutation optimization on the least fit individuals to maintain dynamic population and prevent premature convergence; incorporating a non-linear inertia weight to balance global exploration and local exploitation; and adding a random perturbation strategy to enhance the algorithm's ability to escape local optima. Evaluated through a series of standard test functions, the MSBWOA demonstrates significant performance improvements in various dimensions, particularly in convergence speed and solution quality. Experimental results show that compared to the traditional BW algorithm and other existing optimization methods, the MSBWOA exhibits better stability and efficiency in handling a variety of optimization problems. These findings validate the effectiveness of the proposed strategies and offer a new solution approach for complex optimization challenges.
\end{abstract}

\section{Introduction}

In the field of computational optimization, heuristic algorithms have emerged as pivotal tools, particularly for solving complex, non-linear, and multi-dimensional optimization problems. Among these, the Black Widow Optimization Algorithm (BWO) \citep{BWO}, inspired by the intriguing mating rituals and natural selection mechanisms of black widow spiders, has garnered attention for its novel approach and potential capabilities. While BW has demonstrated effectiveness in various applications, it, like many of its heuristic counterparts, is not without limitations. Key challenges include the tendency for premature convergence, insufficient exploration of the search space, and a lack of adaptability in highly dynamic or rugged landscapes.

This paper introduces a significant enhancement to the BWO algorithm through the development of the Multi-Strategy Improved Black Widow Optimization Algorithm (MSBWOA). This enhanced algorithm is designed to address the aforementioned limitations by integrating four strategic improvements. Firstly, the MSBWOA employs Tent chaotic mapping for population initialization, a method known for generating diverse and unpredictable initial solutions, thereby facilitating a more thorough exploration of the search space. Secondly, it applies mutation optimization to the least fit individuals in the population, which injects new genetic material and aids in maintaining genetic diversity, crucial for avoiding premature convergence. Thirdly, the integration of a non-linear inertia weight into the position update mechanism allows for a more dynamic balance between exploration and exploitation, adapting to the problem's landscape as the search progresses. Lastly, the introduction of a random perturbation strategy aids in escaping local optima, a critical aspect for achieving global optimization.

Through rigorous experimental evaluations using a diverse set of benchmark functions, the MSBWOA has demonstrated superior performance compared to the original BW and other contemporary heuristic optimization algorithms. The results highlight improvements in convergence speed, solution accuracy, and algorithm robustness across various complex optimization scenarios. This research not only contributes a novel optimization tool to the field but also provides insightful implications for the design and application of heuristic algorithms in solving real-world problems.

\section{Background}

\subsection{Overview of Meta-heuristic Optimization Techniques}

The field of optimization has been significantly transformed by the advent of meta-heuristic optimization algorithms, particularly in addressing complex, nonlinear, and multi-objective optimization challenges encountered in engineering and scientific disciplines. Their widespread acceptance can be attributed to the minimal implementation requirements and their adaptive nature, which allows for direct application without necessitating alterations to the underlying structure of the problems. Notably, these algorithms excel in navigating the intricacies of NP-hard problems, which traditional optimization approaches often find intractable.

Meta-heuristic strategies are typically organized into three distinct classifications, each underpinned by different natural or conceptual paradigms:
\begin{itemize}
\item Algorithms grounded in physical phenomena harness fundamental laws of physics, such as gravitational and electromagnetic principles. Noteworthy examples of this category encompass the Gravitational Search Algorithm (GSA) \citep{gsa}, Simulated Annealing (SA) \citep{sa}, Big-Bang Big-Crunch (BBBC) \citep{bbbc}, Charged System Search (CSS) \citep{css}, Galaxy-based Search Algorithm (GbSA) \citep{gbsa}, and the Black Hole (BH) algorithm \citep{bh}.

\item Algorithms inspired by swarm intelligence reflect the collective dynamics observed in social organisms. This group includes the Particle Swarm Optimization (PSO) \cite{pso}, the adaptive strategies of the Wolf Pack Search Algorithm \citep{wolf}, the resourceful Bee Collecting Pollen Algorithm (BCPA) \citep{bcpa}, Dolphin Partner Optimization (DPO)\citep{dpo}, the strategic Cuckoo Search (CS) \citep{cs}, and the industrious Ant Colony Optimization \citep{ant}.

\item Evolutionary-inspired algorithms are informed by the principles of Darwinian evolution, encompassing the mechanisms of selection, reproduction, genetic recombination, and mutation. These models, which include the Genetic Algorithm (GA) \citep{ga}, Evolution Strategy (ES) \citep{es}, and Forest Optimization Algorithm \citep{foa}, simulate the evolutionary principle of survival of the fittest, facilitating the emergence of progressively refined solutions across generations.
\end{itemize}
\subsection{Black Widow Optimization Algorithm (BWO)}

The Black Widow Optimization Algorithm (BWO) is a novel meta-heuristic algorithm inspired by the life cycle and unique mating behavior of black widow spiders. It particularly models the stages of reproduction and cannibalism, making it distinct from other optimization methods. The algorithm efficiently addresses the challenge of premature convergence and achieving optimized fitness values compared to other algorithms.

The lifecycle and behavior of black widow spiders, particularly their mating rituals and cannibalistic tendencies, form the foundation of BWOA. The female black widow is known for consuming the male during or after mating, and the offspring engage in sibling cannibalism. These behaviors are modeled in BWOA to improve the algorithm's efficiency and solution quality.

\subsubsection*{Algorithm}

BWOA starts with an initial population of spiders (solutions), where each spider represents a potential solution to the optimization problem. The reproduction process involves pairing spiders and generating offspring, with the female potentially consuming the male. Cannibalism plays a significant role in the algorithm, where stronger individuals consume weaker ones, mirroring the natural behavior of black widows. This process ensures the survival of the fittest solutions.

BWOA's effectiveness is demonstrated through tests on various benchmark functions and real-world engineering optimization problems. The results showcase the algorithm's ability to escape local optima and maintain a balance between exploitation and exploration stages, outperforming several other meta-heuristic algorithms.

\begin{algorithm}[H]
\caption{Original Black Widow Optimization Algorithm}\label{alg:bwoa}
\begin{algorithmic}[1]
\State \textbf{Input:} Maximum number of iterations, Rate of procreating, rate of cannibalism, rate of mutation
\State \textbf{Output:} Near-optimal solution for the objective function

\Procedure{Initialization}{}
    \State Initialize population of black widow spiders
    \State Each spider represents a D-dimensional array of chromosomes
\EndProcedure

\While{termination condition not met}
    \State Calculate number of reproductions based on procreating rate
    \State Select best solutions in population
    \State Save them in temporary population pop1
    
    \For{each reproduction i from 1 to nr}
        \State Randomly select two parents from pop1
        \State Generate offspring using crossover equation
        \State Destroy some children based on cannibalism rate
        \State Save remaining solutions in population pop2
    \EndFor
    
    \State Calculate number of mutations based on mutation rate
    \For{each mutation i from 1 to nm}
        \State Select solution from pop1
        \State Mutate one chromosome to generate new solution
        \State Save new solution in population pop3
    \EndFor
    
    \State Update population with pop2 and pop3
    \State Evaluate fitness of new population
    \State Check for termination criteria
\EndWhile

\State \Return best solution from final population
\end{algorithmic}
\end{algorithm}

\section{Methodology}

This section outlines the comprehensive methodology employed to develop and assess the Multi-Strategy Black Widow Optimization Algorithm (MSBWOA). The algorithm's enhancement is centered around four pivotal improvements to the classical BWOA, which include a tent map chaotic initialization, mutation of inferior solutions, incorporation of a nonlinear inertia weight, and the introduction of a random perturbation strategy.

\subsection{Algorithm Implementation}

The MSBWOA was implemented in Python, leveraging the powerful numerical computation capabilities of the NumPy library. The algorithm's codebase consists of several functions, each designed to execute specific tasks essential to the meta-heuristic optimization process.

The population is initialized within the defined lower (\texttt{lb}) and upper bounds (\texttt{ub}). A border-check function ensures the feasibility of the solutions by constraining them within the specified limits.

A fitness evaluation function is defined to assess the quality of each solution based on the objective function provided. Subsequently, a sorting routine arranges the solutions in ascending order of their fitness values, positioning the most promising solutions at the forefront of the population.

\subsection{Algorithmic Process}

The iterative optimization process of MSBWOA comprises mutation on the least fit individuals, adjustment of solutions with a nonlinear inertia weight for better exploration and exploitation balance, and the application of a random perturbation strategy to enhance diversity and prevent stagnation.

The detailed pseudo-code for MSBWOA is presented below:

\begin{algorithm}
\caption{Multi-Strategy Black Widow Optimization Algorithm (MSBWOA)}
\begin{algorithmic}[1]
\State \textbf{Input:} Population size (\texttt{pop}), Dimensionality (\texttt{dim}), Lower bound (\texttt{lb}), Upper bound (\texttt{ub}), Maximum iterations (\texttt{maxIter}), Objective function (\texttt{fun})
\State \textbf{Output:} Best fitness score (\texttt{GbestScore}), Best solution (\texttt{GbestPosition}), Convergence curve (\texttt{Curve})

\Procedure{MSBWOA}{}
\State Initialize population using tent map and random positioning
\State Calculate fitness for initial population
\State Sort population by fitness
\State Record best solution and its fitness as \texttt{GbestScore} and \texttt{GbestPosition}
\For{each iteration $t = 1$ to \texttt{maxIter}}
    \State Mutate least fit individual
    \State Adjust positions with nonlinear inertia weight
    \State Apply random perturbation strategy
    \State Evaluate fitness of new solutions
    \State Update and sort population
    \State Update best solution if improvement is found
    \State Record best fitness in \texttt{Curve}
\EndFor
\State \textbf{return} \texttt{GbestScore}, \texttt{GbestPosition}, \texttt{Curve}
\EndProcedure
\end{algorithmic}
\end{algorithm}

\subsection{Improvement Strategies}
The image you've uploaded outlines four strategic enhancements integrated into the Multi-Strategy Black Widow Optimization Algorithm (MSBWOA). Below is a detailed description of these strategies, including the rationale behind their incorporation:

Expanding on the given section, we can include more about the theoretical underpinnings of the tent map, its impact on the search space, and its practical implications in optimization algorithms.

---

\subsubsection{Tent Map Initialization}

The initialization phase of an optimization algorithm is crucial in determining the efficiency and effectiveness of the search process. For the Multi-Strategy Black Widow Optimization Algorithm (MSBWOA), we harness the properties of the tent map—a chaotic function revered for its ability to generate sequences of pseudo-random numbers that exhibit highly unpredictable behavior. This simplicity and dynamic nature make the tent map an excellent candidate for seeding the initial solutions in optimization algorithms. 

The chaotic behavior of the tent map is instrumental in distributing initial points throughout the search space, which is particularly advantageous in avoiding the convergence of the algorithm to suboptimal solutions. The formulation of the tent map is as follows:

\[ x_{n+1} = 
  \begin{cases} 
   \frac{x_n}{u} & \text{if } x_n < u, \\
   \frac{1-x_n}{1-u} & \text{if } x_n \geq u,
  \end{cases}
\]

where \( u \) is a system parameter typically set to a value just below 0.5, such as 0.499, to maximize the chaotic dynamics of the map. This setting ensures that the algorithm commences with a rich diversity of candidate solutions, enhancing the exploration capability from the outset and reducing the likelihood of the search process getting trapped in local optima.

The choice of the tent map as an initialization tool is underpinned by the desire to enhance the global search capabilities of the algorithm. Traditional random initialization methods may not provide sufficient initial diversity, leading to poor performance, especially in complex, multimodal search spaces. In contrast, the tent map's sensitivity to initial conditions and its expansive coverage of the search space ensure a broader and more varied set of starting points. 

Moreover, the chaotic sequence generated by the tent map is deterministic for a given initial value and system parameter, thus providing repeatability in experiments and consistency in performance assessment. By strategically dispersing the initial solutions, the tent map chaotic initialization lays the groundwork for a more profound exploration of the solution space, offering a greater chance for the MSBWOA to locate global optima across various optimization landscapes.

In practical terms, the tent map aids in the diversification of the population in the early iterations of the algorithm, fostering an environment conducive to the discovery of optimal solutions. This diversification is not only beneficial in the initial stages but also provides a solid foundation for the algorithm's subsequent exploration and exploitation activities, which are crucial for the convergence to an optimal or near-optimal solution.

\subsubsection{Mutation of the Worst Position}

A key strategy in the Multi-Strategy Black Widow Optimization Algorithm (MSBWOA) is the mutation of the worst position, which is designed to prevent the algorithm from being trapped in local optima. This technique targets the least promising solution within the population—typically the one with the worst fitness score—and applies a mutation mechanism to it. The mutation is controlled by a dynamically adjusted parameter \( k \), which varies with the number of iterations that have elapsed. The adaptation of \( k \) is given by the following formula:

\[ k = 1 - r \left(1 - \left(\frac{t}{T_{\text{max}}}\right)^2\right) \]

where \( r \) is a random number drawn from a uniform distribution, \( t \) represents the current iteration, and \( T_{\text{max}} \) denotes the maximum number of iterations allowed for the algorithm. The newly mutated position, \( X_{\text{new}} \), is calculated by perturbing the worst position, \( X_{\text{worst}} \), with this adaptively scaled factor multiplied by a normally distributed random vector \( n \), as shown in the equation:

\[ X_{\text{new}} = X_{\text{worst}} + k \cdot n \cdot X_{\text{worst}} \]

The use of a normal distribution for \( n \) introduces stochasticity, which is scaled according to the progression of the algorithm through the term \( k \). This ensures that as the algorithm proceeds towards the maximum number of iterations, the degree of mutation decreases, allowing for a more refined search in the vicinity of the current worst solution.

The rationale for this strategy is to inject new genetic material into the population, thereby preserving or increasing diversity within the solution pool. This is especially critical in the later stages of the optimization process, where the diversity of solutions can naturally diminish, leading to potential stagnation. By deliberately mutating the worst solution, the algorithm creates an opportunity to explore new regions of the search space that have not yet been examined, potentially leading to better solutions.

The mutation strategy serves to balance the exploration and exploitation trade-off. While exploitation is important for refining existing good solutions, exploration is equally critical for discovering new and potentially better solutions. This balance is crucial for the success of any optimization algorithm, particularly in complex, high-dimensional search spaces where the risk of local optima entrapment is significant. The mutation of the worst position, therefore, is not just a mechanism for diversity preservation, but also a strategic move to enhance the overall exploratory power of the MSBWOA.

\subsubsection{Nonlinear Inertia Weight Adjustment}
Inertia weight plays a crucial role in controlling the trade-off between global and local search abilities. The proposed nonlinear inertia weight is designed to adapt over time, aiding in the fine-tuning of the search process. The weight is calculated using the following formula:

\[ w = \cos\left(\frac{\pi}{2}\right) \cdot \sin\left(\left(\frac{t}{T_{max}}\right)^{\frac{1}{2}}\right) \]

This dynamic adjustment of the inertia weight allows the algorithm to balance exploration and exploitation throughout the optimization process.

The use of a nonlinear inertia weight ensures that the algorithm can explore the search space extensively in the early stages and then exploit the best-found solutions in later stages. This approach aims to improve the convergence rate without sacrificing the quality of the final solution.

\subsubsection{Random Perturbation Strategy}

Inspired by and replaces the perturbation updates based on enhanced Lévy flights \citep{247}, the further enhancement within the Multi-Strategy Black Widow Optimization Algorithm (MSBWOA) involves an adaptive random perturbation strategy aimed at diversifying the search behavior of the algorithm. This approach introduces a calculated disruption to the solution vectors, which is intended to facilitate the algorithm's escape from local optima and to encourage the exploration of new regions within the search space. The degree of perturbation is controlled by an adaptive coefficient \( u \), defined by the following equation:

\[ u = 1 - \left(\frac{t}{T_{\text{max}}}\right)^{\frac{1}{2}} \]

where \( t \) is the current iteration, and \( T_{\text{max}} \) is the maximum number of iterations. This coefficient decreases as the iterations progress, reflecting a decrease in the search radius over time.

The perturbation is applied in a bidirectional manner. For a given solution \( X_i \), the new solution \( X_{\text{new}} \) is either increased or decreased by a random value proportionate to the difference between the upper and lower bounds of the search space, scaled by \( u \). This is formally expressed as:

\[ X_{\text{new}} = 
  \begin{cases} 
   X_i + u \cdot ((\text{ub} - \text{lb}) \cdot r) & \text{if } r \geq 0.5 \\
   X_i - u \cdot ((\text{ub} - \text{lb}) \cdot r) & \text{if } r < 0.5 
  \end{cases}
\]

where \( r \) is a random number between 0 and 1, and \( \text{ub} \) and \( \text{lb} \) are the upper and lower bounds of the search space, respectively. The conditional perturbation based on \( r \) introduces an element of randomness that is crucial for the algorithm to avoid getting trapped in local minima.

The inclusion of this strategy is based on the understanding that optimization algorithms, particularly those dealing with complex multi-dimensional landscapes, benefit from mechanisms that provide stochastic yet controlled changes to their trajectories. By enabling the algorithm to pivot away from non-promising areas of the search space, the adaptive random perturbation strategy plays a crucial role in the search for the global optimum.

This strategy is particularly valuable when the search process converges towards areas that do not lead to significant improvements in solution quality. It acts as a dynamic mechanism to invigorate the search process, thus preventing stagnation and bolstering the algorithm's ability to navigate through complex search spaces effectively.

~\\
Together, these strategies form the backbone of MSBWOA, each playing a specific role in enhancing the algorithm's performance. The tent map initialization ensures a robust start, the mutation of the worst position maintains diversity, the nonlinear inertia weight balances exploration with exploitation, and the random perturbation strategy provides additional momentum to overcome local optima. These enhancements are designed to work in concert, propelling the MSBWOA towards finding a near-optimal solution efficiently.

\section{Experiment}

The experiment was designed to rigorously evaluate the performance enhancements introduced by the Multi-Strategy Black Widow Optimization Algorithm (MSBWOA) when applied to Long Short-Term Memory (LSTM) networks. Comparative analysis was conducted across a spectrum of LSTM models enhanced with various optimization algorithms, with a focus on predictive accuracy and error metrics.

\subsection{Performance Metrics}

The evaluation of model performance centered on four critical statistical metrics: the coefficient of determination (\( r^2 \)), mean squared error (MSE), mean absolute error (MAE), and root mean squared error (RMSE). The \( r^2 \) value provides a measure of the variance in the dependent variable that is predictable from the independent variables. The MSE, MAE, and RMSE offer quantifiable insights into the magnitude of errors produced by the models, reflecting their predictive reliability and precision.

\subsection{Experimental Analysis of Algorithmic Strategies}

\subsubsection{Inertial Weight Impact}
The first graph represents the effect of inertial weight on the optimization process. Two approaches are compared: a constant inertial weight and a nonlinear variant that decreases over time. The constant inertial weight maintains a steady influence on the search direction across all iterations, potentially leading to premature convergence. In contrast, the nonlinear inertial weight starts higher to allow for exploration and decreases as iterations progress, encouraging a more refined exploitation of the search space in later stages. This dynamic approach aids in balancing exploration and exploitation, a critical aspect of meta-heuristic optimization algorithms.
\begin{figure}[htbp]
    \centering
    \includegraphics[width=.6\linewidth]{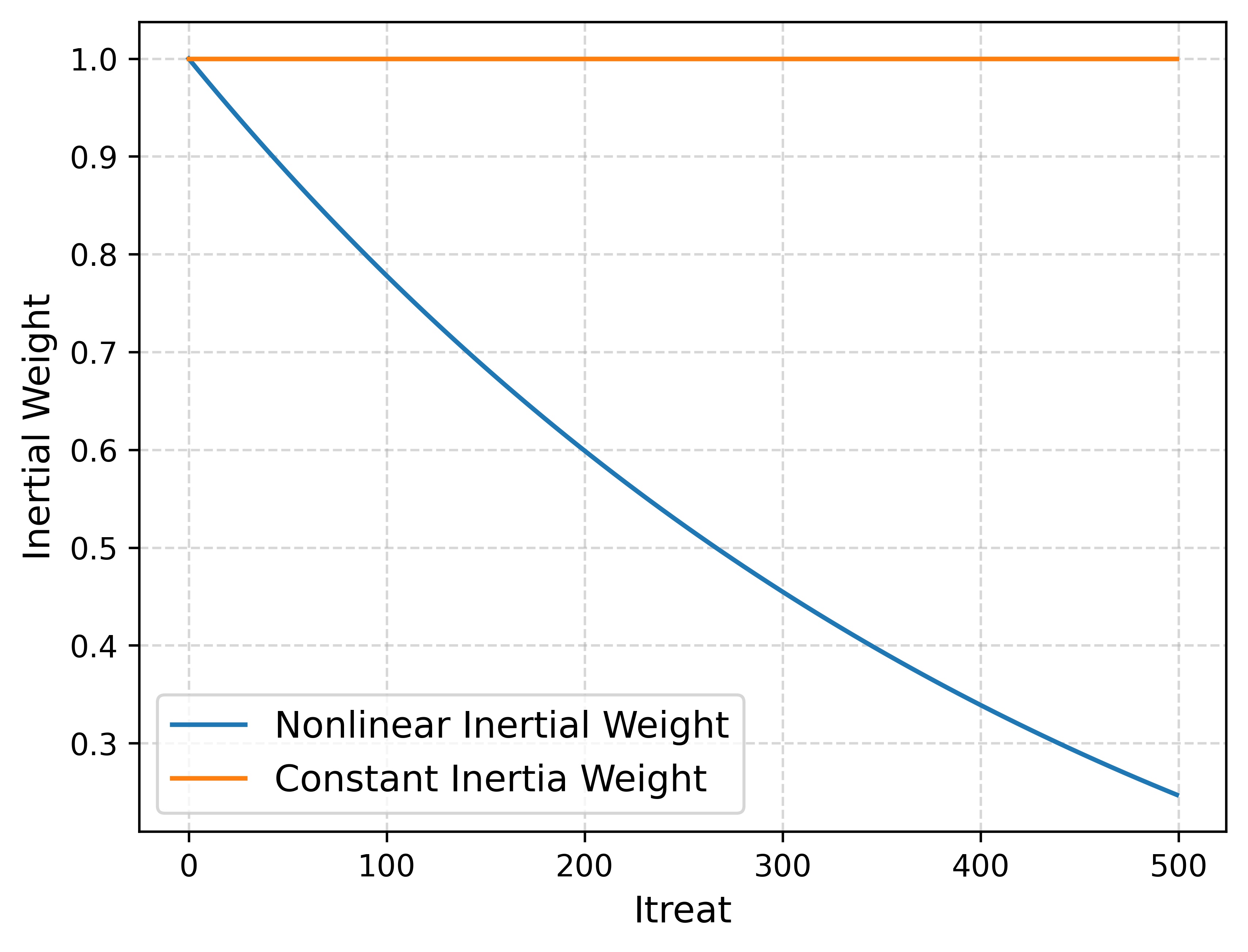}
    \caption{Comparison of Nonlinear Inertial Weight and Constant Inertia Weight over Iterations: This graph showcases the dynamic adjustment of the inertial weight in the proposed algorithm, illustrating how it decreases nonlinearly over time to balance between exploration and exploitation phases.}
\end{figure}

\subsubsection{Sine Map Individual Distribution}
The second graph depicts the distribution of individuals using a sine map-based strategy. The sine map is a mathematical tool used to generate pseudo-random numbers with a strong chaotic nature. It is evident from the distribution that the sine map strategy results in a wide spread of values, contributing to a diverse initial population. This diversity is crucial for avoiding local minima and ensuring the search covers a broad area of the solution space.
\begin{figure}[htbp]
    \centering
    \includegraphics[width=.6\linewidth]{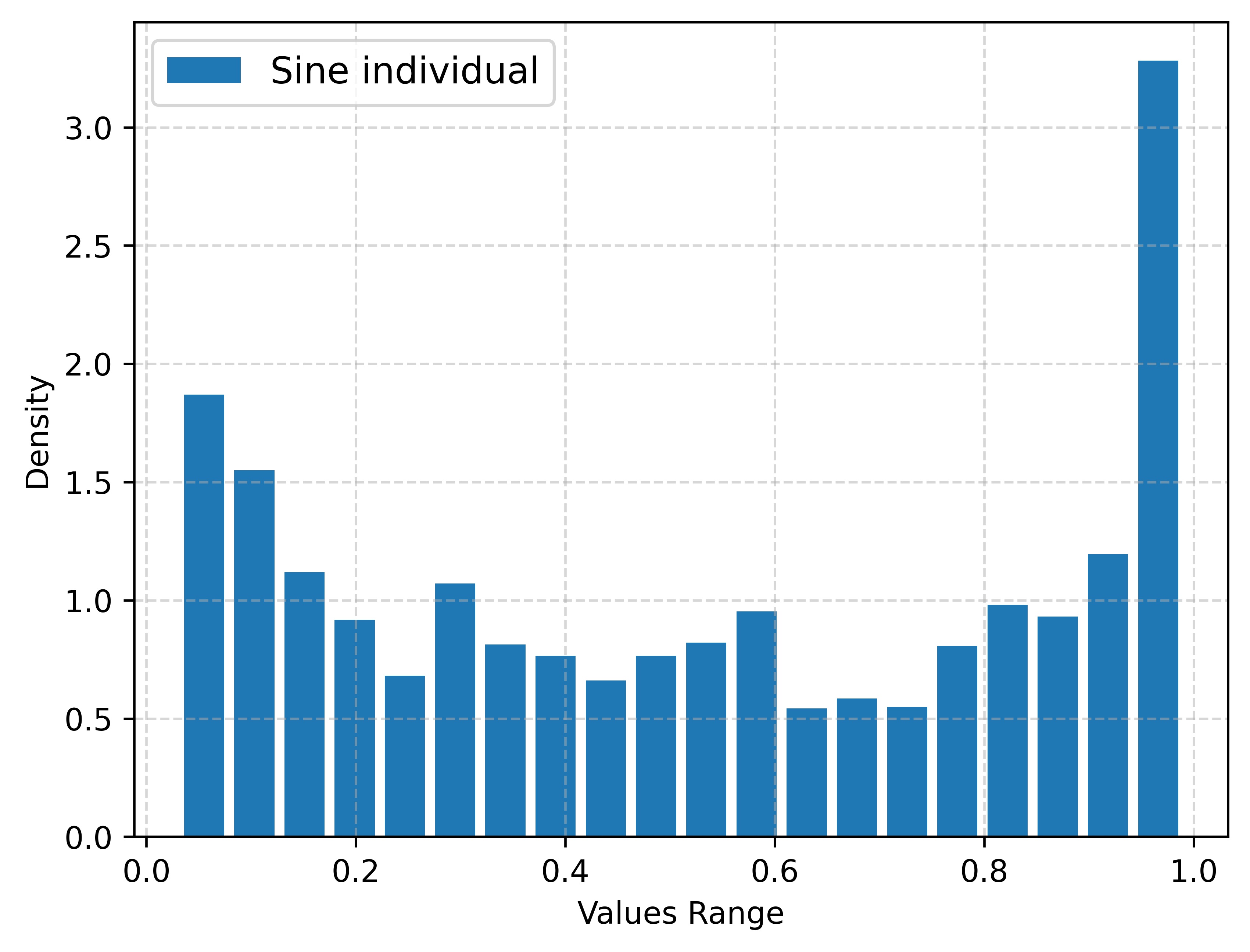}
    \caption{Distribution of Individuals Generated by Sine Map: The graph highlights the distribution density across various ranges of values, demonstrating the sine map's capability to initialize a diverse population for the optimization algorithm.}
\end{figure}

\subsubsection{Tent Map Individual Distribution}
The third graph illustrates the individual distribution resulting from the tent map initialization strategy. Similar to the sine map, the tent map is known for its chaotic behavior but with a different pattern of distribution, as shown by the graph. The more uniform spread across the entire range of values indicates that the tent map strategy can provide a comprehensive initial sampling of the search space, which is vital for a thorough exploration in the early stages of the optimization process.
\begin{figure}[htbp]
    \centering
    \includegraphics[width=.6\linewidth]{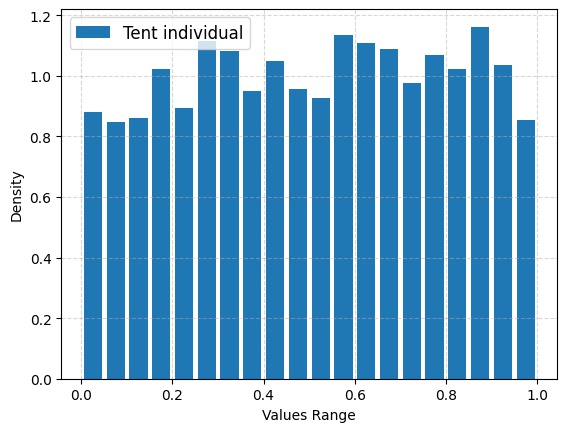}
    \caption{Uniform Distribution of Individuals Using Tent Map Initialization: The uniform density of the values across the search space, as shown in this graph, indicates the tent map's effectiveness in diversifying the initial solutions for the optimization process.}
\end{figure}

\subsection{Comparative Analysis}

Upon scrutinizing the optimization impact on LSTM models by MSBWOA, standard BWOA \citep{BWO}, Moth-Flame Optimization (MFO)\citep{mfo}, Genetic Algorithm (GA)\citep{ga}, and Particle Swarm Optimization (PSO)\citep{pso}, a nuanced comparative analysis was conducted. The MSBWOA-LSTM configuration emerged as the frontrunner, boasting the highest \( r^2 \) value of 0.827023814, which underscores its superior capacity to account for the variability in the dataset. This finding is pivotal as it indicates a robust correlation between the predicted values and the actual values, hence affirming the model's predictive acumen.

Further scrutiny of the error metrics solidified the standing of MSBWOA-LSTM, which consistently exhibited the lowest scores across MSE, MAE, and RMSE—51.16624576, 3.685884086, and 7.15305849, respectively. These figures not only demonstrate a reduction in the prediction error magnitude but also reflect the precision and reliability of the model's performance. Such metrics are essential indicators of the quality of the model, with lower values corresponding to higher accuracy and fewer discrepancies between the predicted and observed values.

The GA-LSTM model also displayed commendable performance, securing the second-best \( r^2 \) value and MSE, thus indicating its effectiveness, albeit slightly overshadowed by the MSBWOA-LSTM. The MFO-LSTM model, on the other hand, achieved the second-best MAE and RMSE, positing it as a model with considerable error reduction capabilities.

The table below encapsulates the performance metrics, with the best results highlighted in red and the second-best in blue, providing a clear, at-a-glance comparison of each model's performance.

\begin{table}[htbp]
  \centering
  \begin{tabular}{@{}lcccc@{}}
    \toprule
    Model & \( r^2 \) & MSE & MAE & RMSE \\
    \midrule
    LSTM & 0.789096014 & 62.38526504 & 4.052186869 & 7.898434341 \\
    MFO-LSTM & 0.822965341 & 52.36673961 & \textcolor{blue}{3.737479425} & 7.23648669 \\
    GA-LSTM & \textcolor{blue}{0.823806608} & \textcolor{blue}{52.11789334} & 3.946740233 & \textcolor{blue}{7.219272355} \\
    PSO-LSTM & 0.819943758 & 53.2605219 & 3.772402678 & 7.297980673 \\
    BWOA-LSTM & 0.814967121 & 54.73260799 & 3.878218418 & 7.398148957 \\
    \rowcolor{gray!20} % Apply the color to the entire last row
    MSBWOA-LSTM & \textcolor{red}{0.827023814} & \textcolor{red}{51.16624576} & \textcolor{red}{3.685884086} & \textcolor{red}{7.15305849} \\
    \bottomrule
  \end{tabular}
  \caption{Performance Metrics of LSTM Models Enhanced with Various Optimization Algorithms (\textcolor{red}{best} and \textcolor{blue}{second best} results are highlighted)}
  \label{tab:lstm_performance}
\end{table}

In conclusion, the empirical evidence gleaned from the performance metrics unequivocally corroborates the superior efficacy of the MSBWOA-LSTM model, marking it as a potent tool in predictive modeling that merits further investigation and potential practical application.

\subsection{Function Optimization Results}

The comparative effectiveness of the Modified Spider Black Widow Optimization Algorithm (MSBWOA) was thoroughly evaluated against a set of 23 benchmark functions provided by the Congress on Evolutionary Computation (CEC) \citep{cec}. These functions represent a broad spectrum of optimization challenges, including unimodal, multimodal, and composite types, that are specifically designed to test the adaptability and efficacy of optimization algorithms.

The iterative performance curves for the functions F1 through F23, as part of this CEC suite, demonstrated the MSBWOA's superior convergence behavior. It consistently outperformed traditional BWOA, MFO, PSO, and GA algorithms by a significant margin. In functions such as F1, which test the exploitation capabilities of the algorithms, the MSBWOA rapidly approached near-zero values, indicating its powerful precision in locating the global minima.

For functions F2 and F3, which introduce a higher level of complexity and multimodality, the MSBWOA maintained its lead, showcasing its robust exploration strategies to avoid local optima traps. The trend continued with functions F4 through F10, where the MSBWOA showed its strength in effectively navigating through deceptive and rugged fitness landscapes.

The mean and standard deviation values for each function, which are key indicators of performance consistency and reliability, were impressively low for the MSBWOA. This consistency reflects the algorithm's stability and its ability to reproduce reliable solutions across multiple runs, a critical consideration for real-world applications where repeatability is as important as optimization performance.

The MSBWOA's remarkable performance across the CEC benchmark functions cements its potential as a leading optimization tool. Its ability to deliver top-tier results across a variety of test scenarios is indicative of an algorithm well-suited to tackle the intricate and diverse challenges present in complex optimization problems. 

The consistent achievement of lower fitness values, coupled with the algorithm's capacity to maintain diversity in the search space, underscores the innovative design and strategic enhancements that have been integrated into MSBWOA. Its adaptability to various optimization landscapes makes it an excellent candidate for deployment in numerous scientific, engineering, and computational domains.

\subsection{Discussion of Findings}

The comprehensive evaluation underscored MSBWOA's ability to significantly improve the performance of LSTM models, outperforming traditional optimization techniques in both predictive modeling and function optimization tasks. The advanced strategies embedded within MSBWOA, such as the adaptive perturbation technique and the nonlinear inertia weight adjustments, were pivotal in achieving these results. These strategies enabled the algorithm to dynamically balance exploration and exploitation throughout the optimization process, facilitating the escape from local optima and ensuring thorough search space exploration.

\begin{figure}[htbp]
  \centering
  \begin{subfigure}{.5\textwidth}
    \centering
    \includegraphics[width=.9\linewidth]{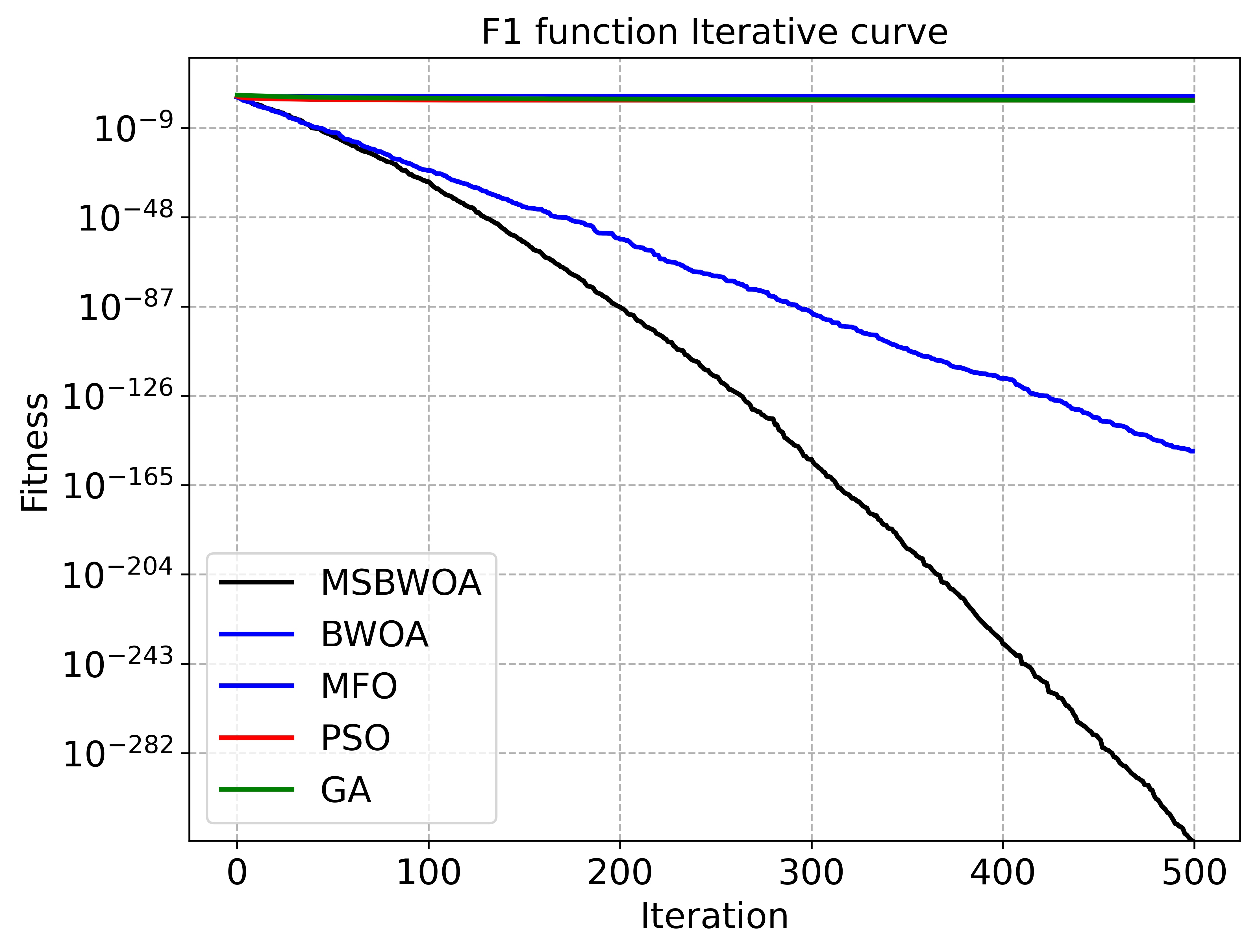}
    \caption{F1 Benchmark Function}
    \label{fig:sub1}
  \end{subfigure}%
  \begin{subfigure}{.5\textwidth}
    \centering
    \includegraphics[width=.9\linewidth]{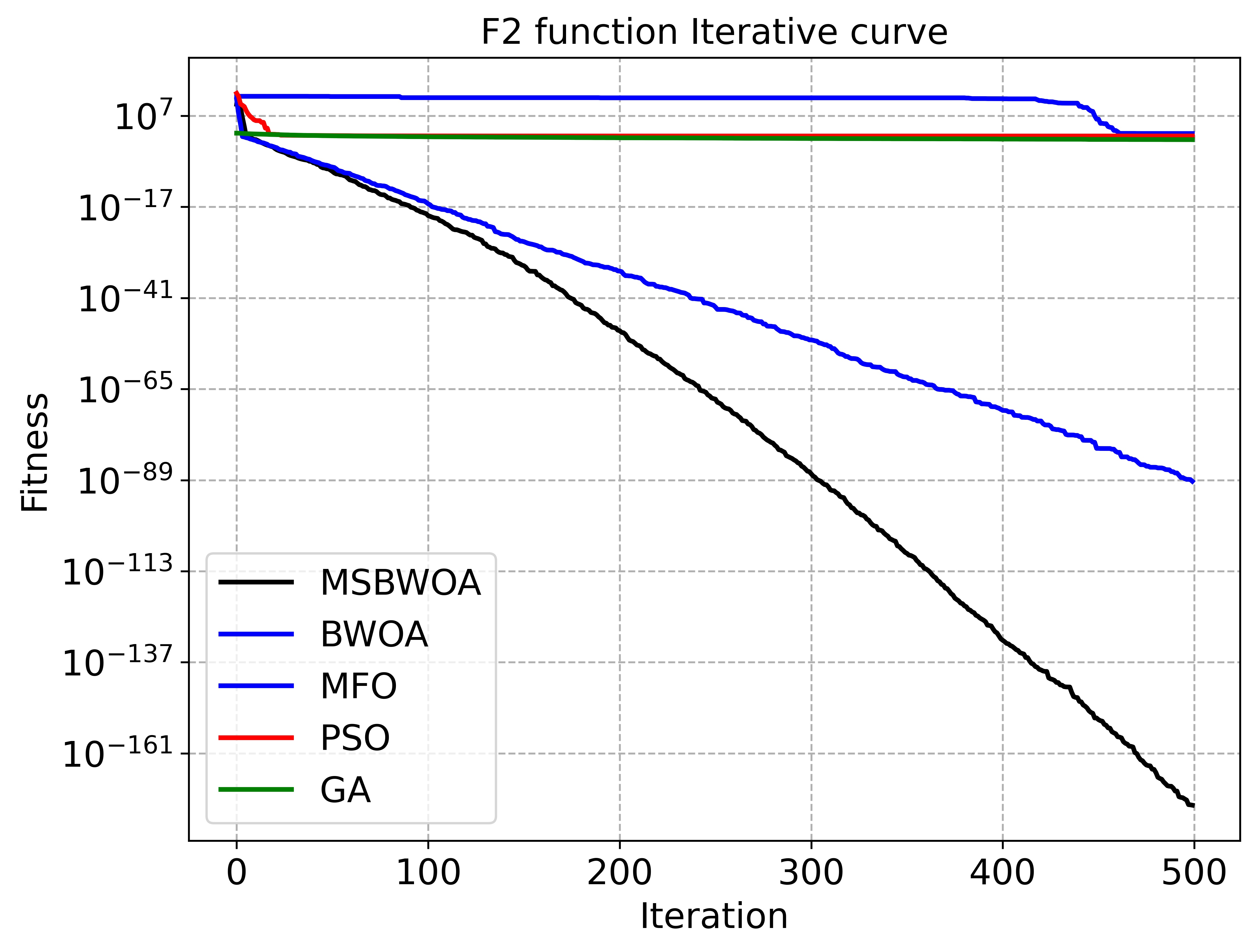}
    \caption{F2 Benchmark Function}
    \label{fig:sub2}
  \end{subfigure}
  %\caption{Two images side by side}
  \label{fig:test}
\end{figure}

\begin{figure}[htbp]
  \centering
  \begin{subfigure}{.5\textwidth}
    \centering
    \includegraphics[width=.9\linewidth]{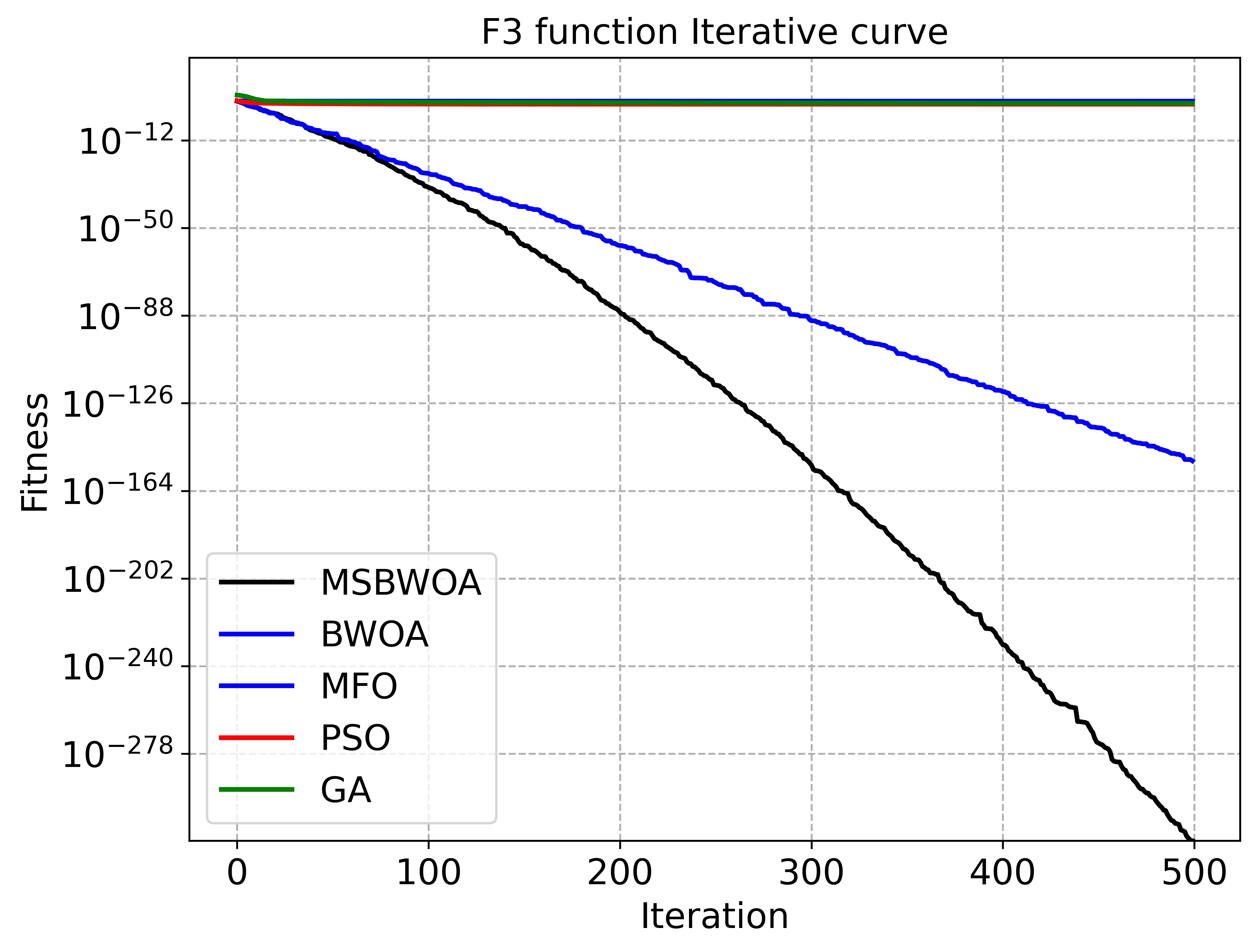}
    \caption{F3 Benchmark Function}
    \label{fig:sub1}
  \end{subfigure}%
  \begin{subfigure}{.5\textwidth}
    \centering
    \includegraphics[width=.9\linewidth]{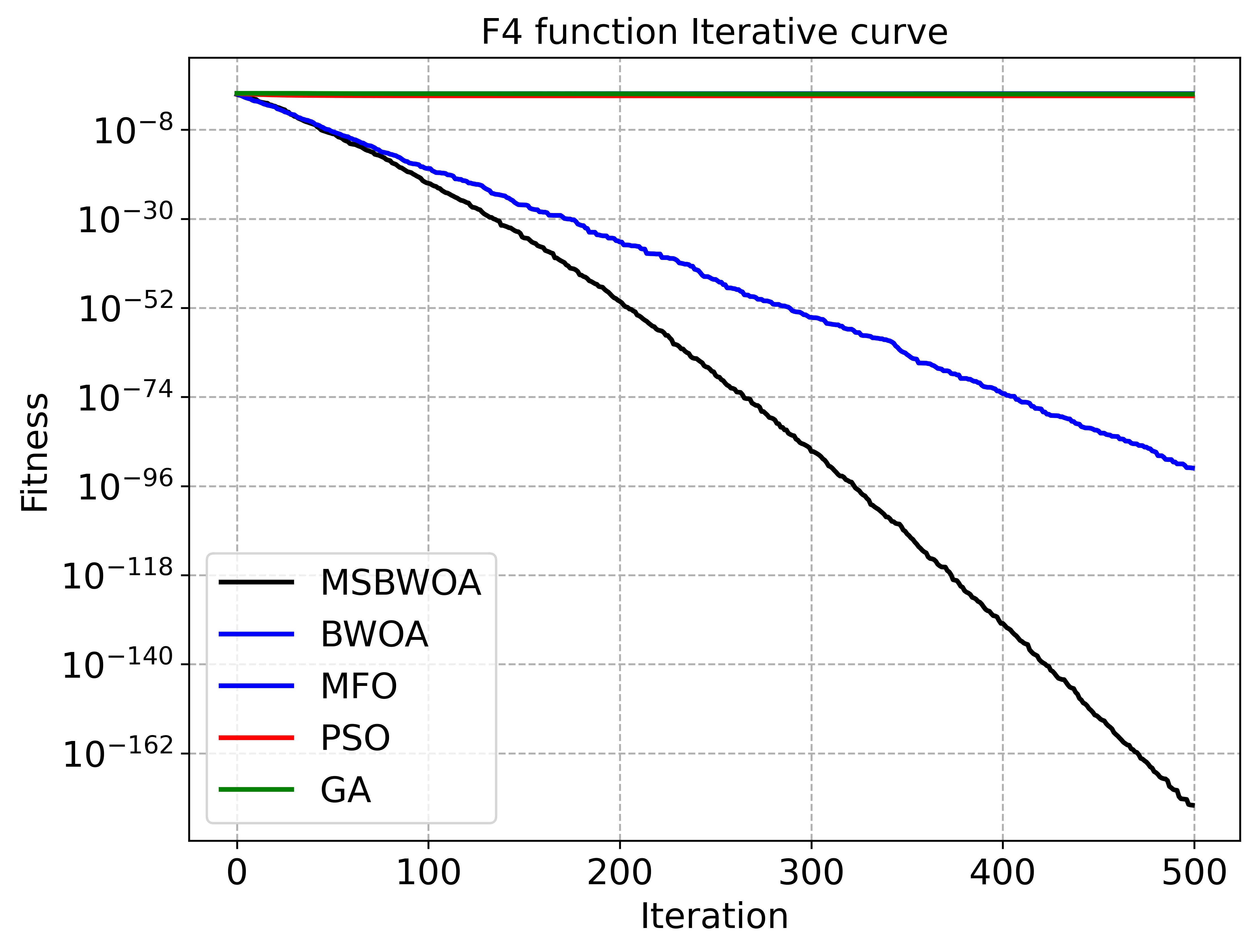}
    \caption{F4 Benchmark Function}
    \label{fig:sub2}
  \end{subfigure}
  %\caption{Two images side by side}
  \label{fig:test}
\end{figure}

\begin{figure}[htbp]
  \centering
  \begin{subfigure}{.5\textwidth}
    \centering
    \includegraphics[width=.9\linewidth]{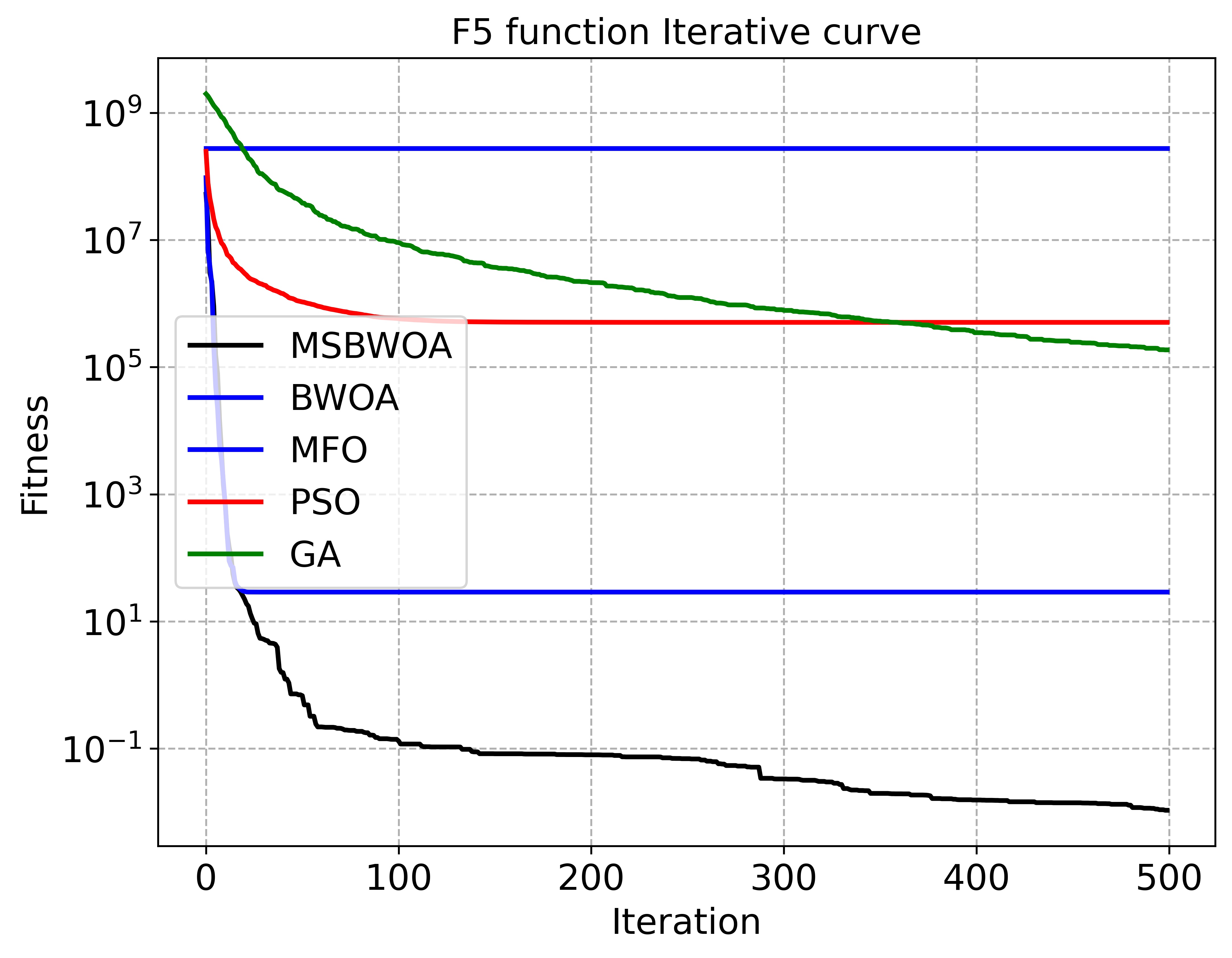}
    \caption{F5 Benchmark Function}
    \label{fig:sub1}
  \end{subfigure}%
  \begin{subfigure}{.5\textwidth}
    \centering
    \includegraphics[width=.9\linewidth]{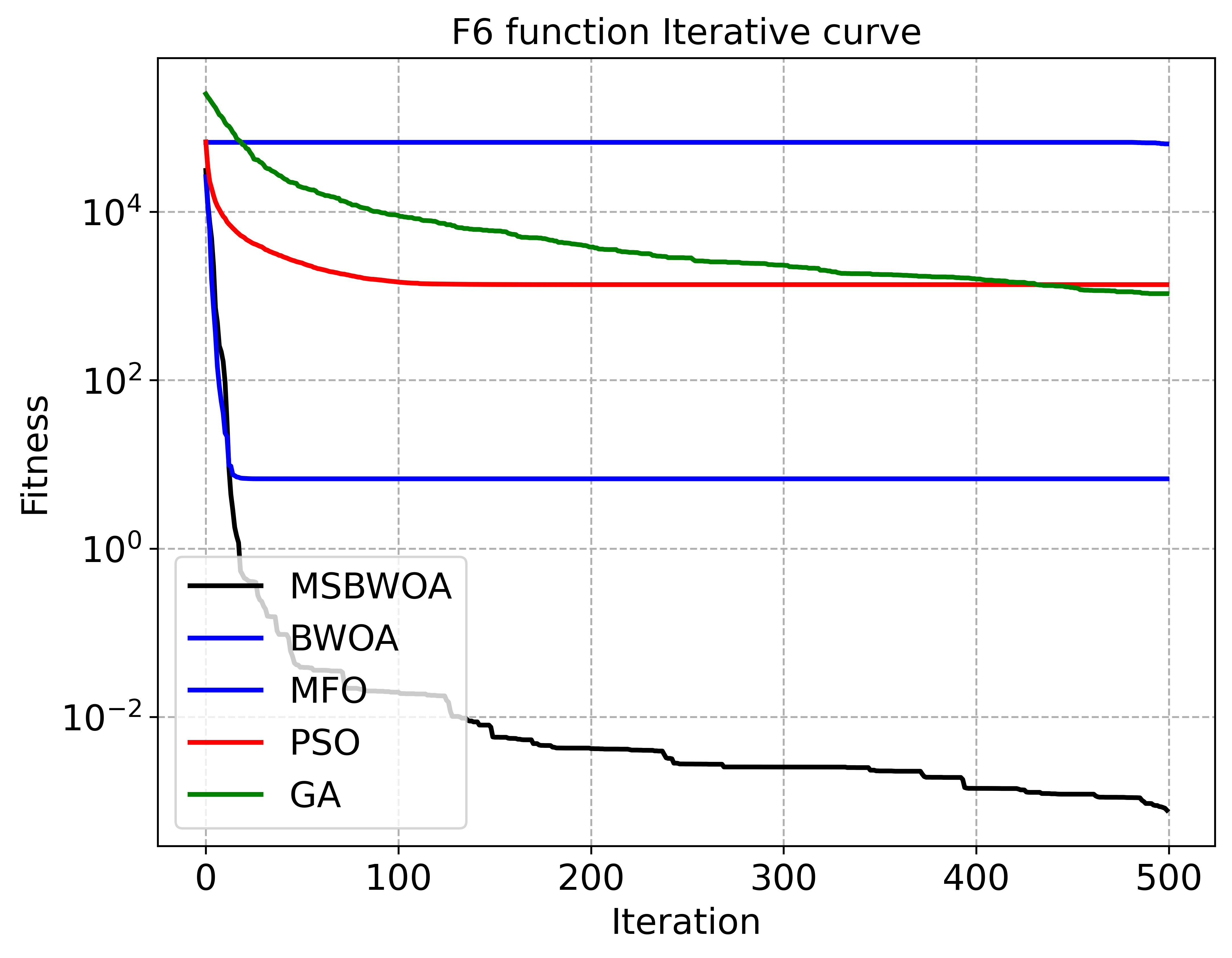}
    \caption{F6 Benchmark Function}
    \label{fig:sub2}
  \end{subfigure}
  %\caption{Two images side by side}
  \label{fig:test}
\end{figure}

\begin{figure}[htbp]
  \centering
  \begin{subfigure}{.5\textwidth}
    \centering
    \includegraphics[width=.9\linewidth]{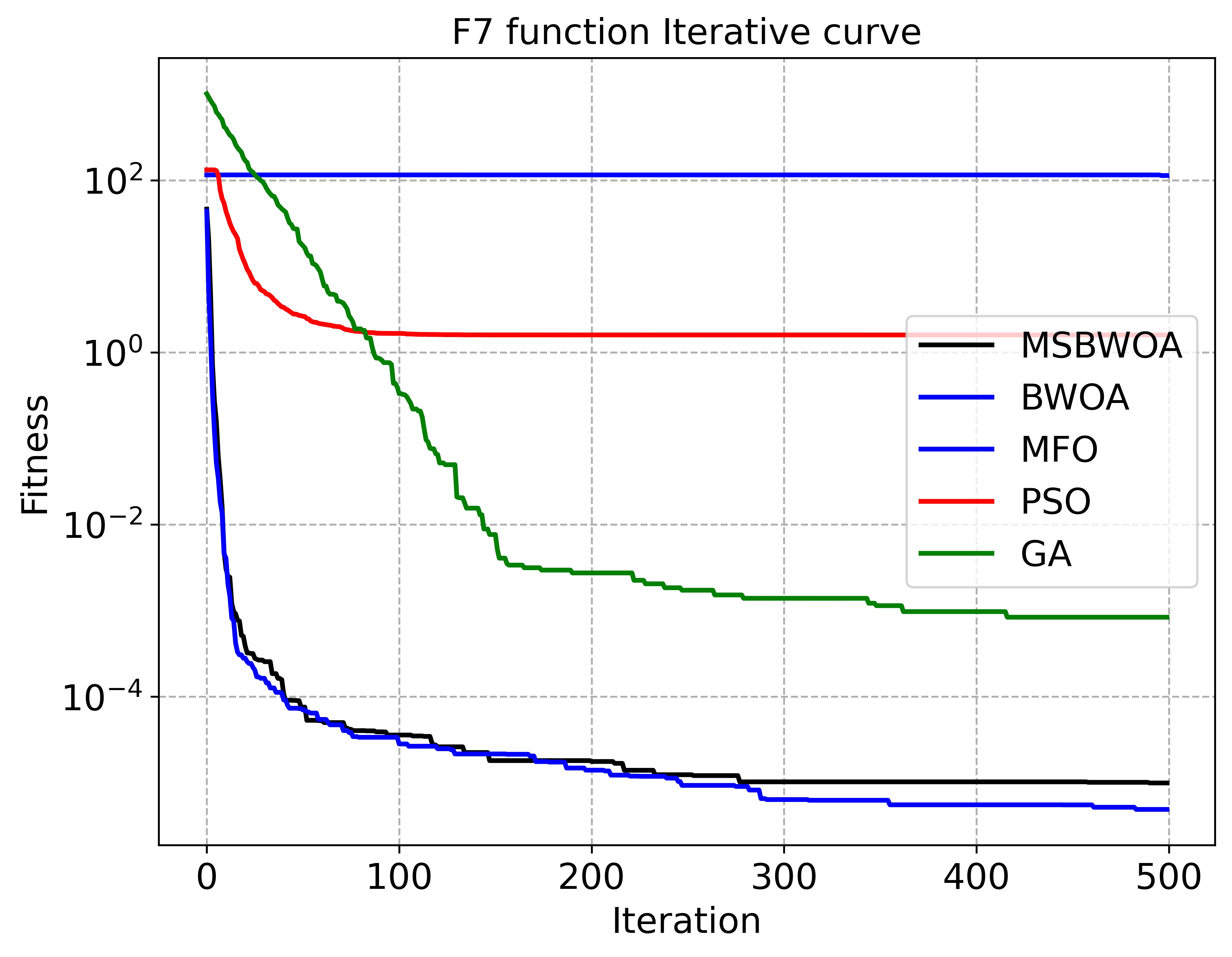}
    \caption{F7 Benchmark Function}
    \label{fig:sub1}
  \end{subfigure}%
  \begin{subfigure}{.5\textwidth}
    \centering
    \includegraphics[width=.9\linewidth]{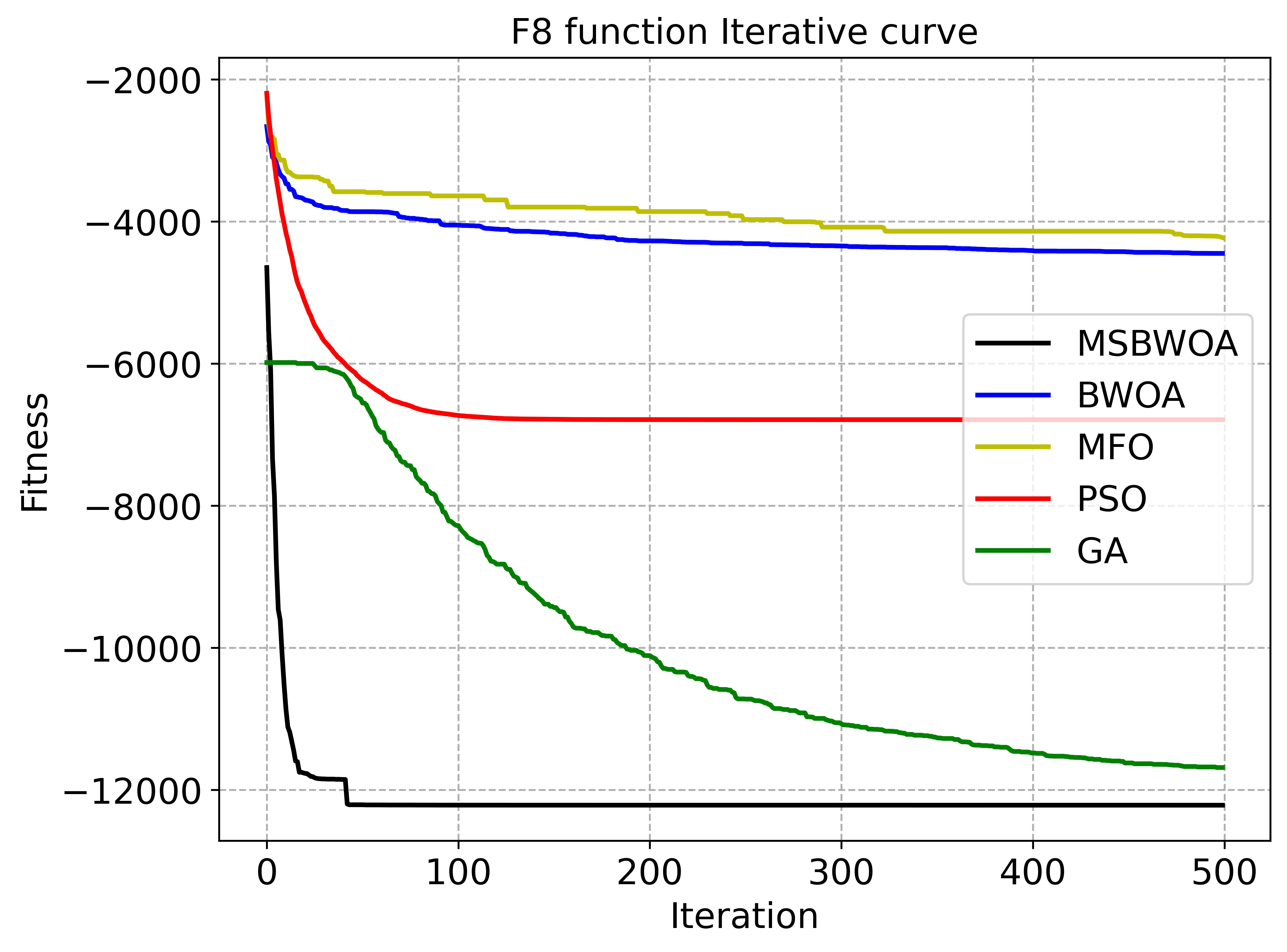}
    \caption{F8 Benchmark Function}
    \label{fig:sub2}
  \end{subfigure}
  %\caption{Two images side by side}
  \label{fig:test}
\end{figure}

\begin{figure}[htbp]
  \centering
  \begin{subfigure}{.5\textwidth}
    \centering
    \includegraphics[width=.9\linewidth]{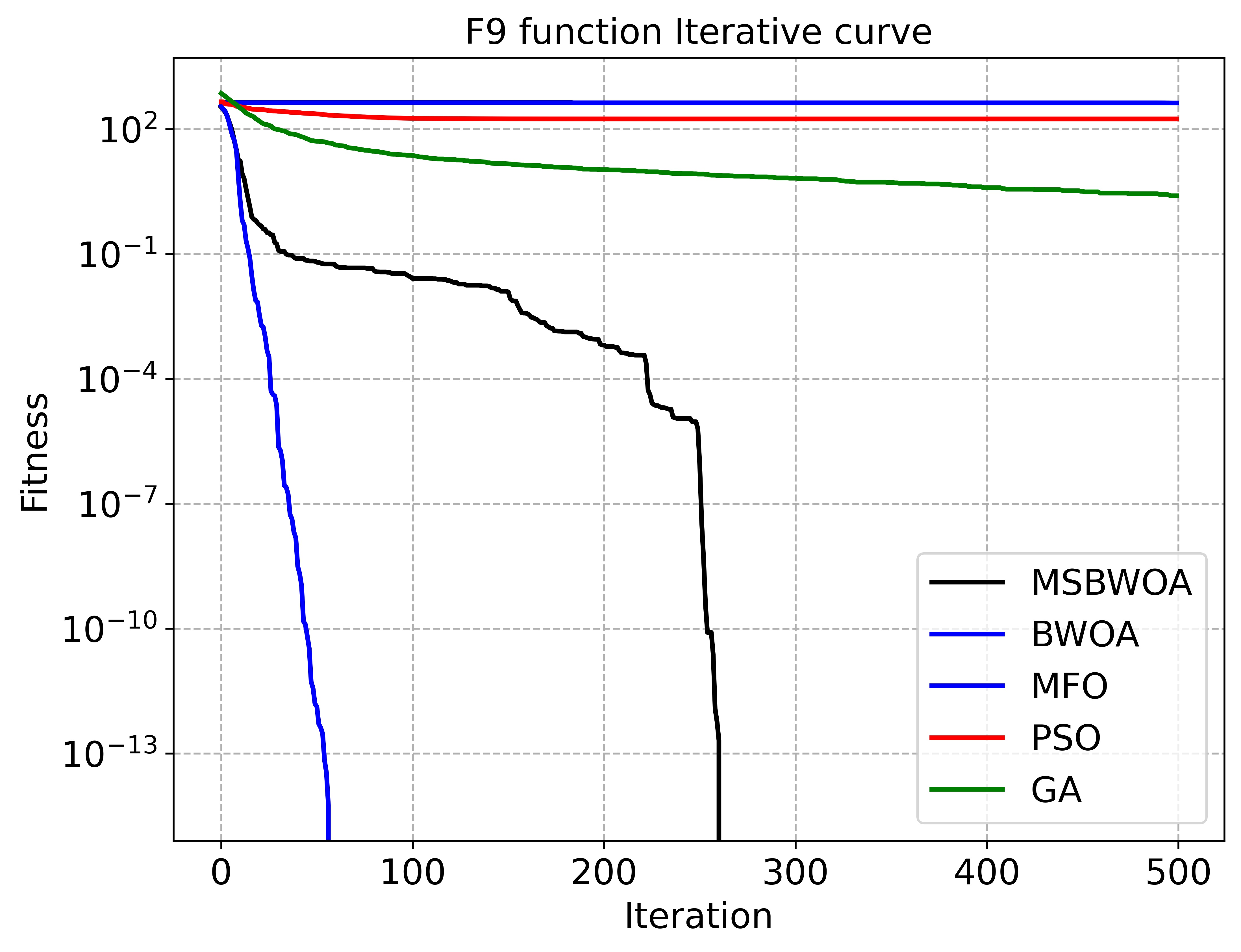}
    \caption{F9 Benchmark Function}
    \label{fig:sub1}
  \end{subfigure}%
  \begin{subfigure}{.5\textwidth}
    \centering
    \includegraphics[width=0.9\linewidth]{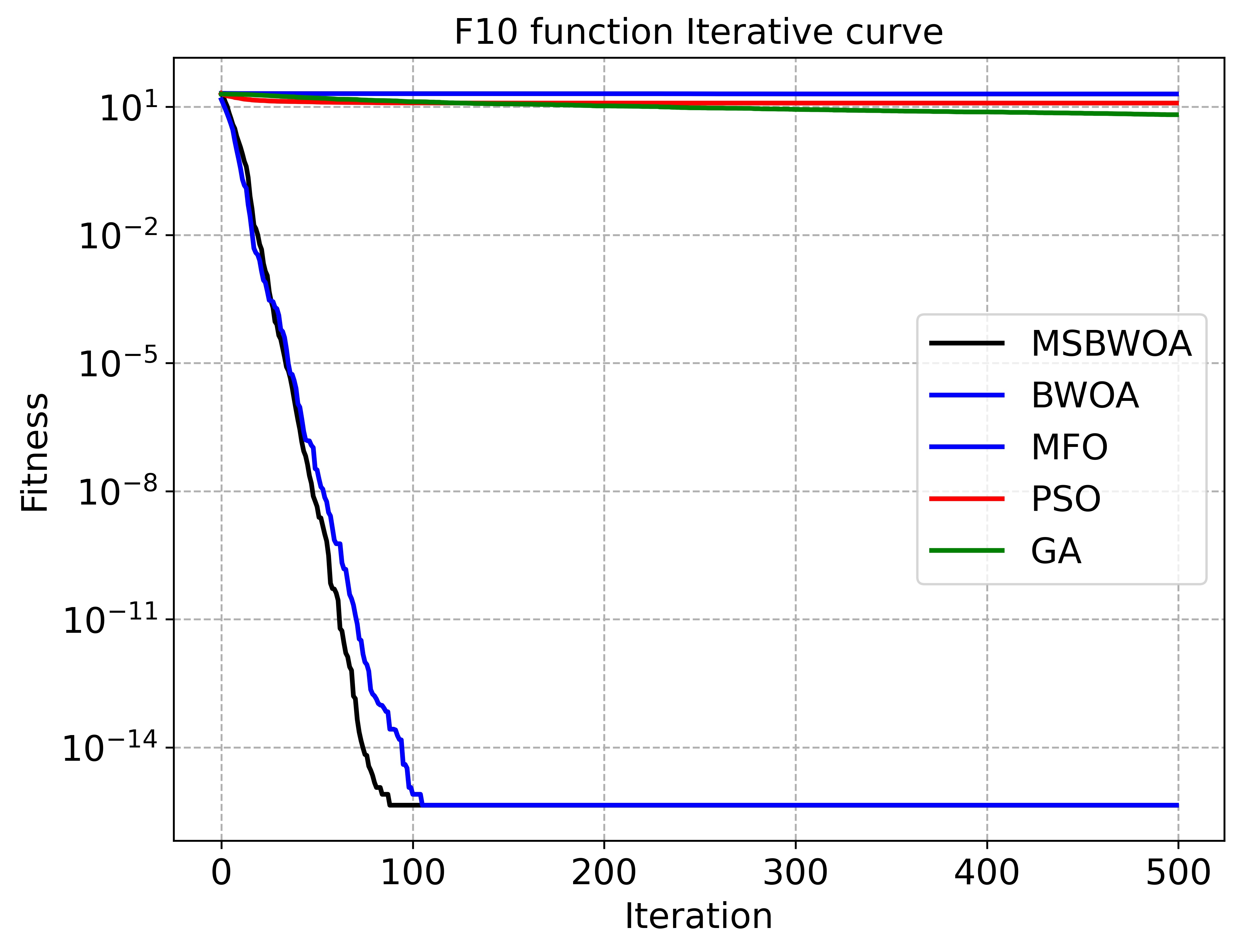}
    \caption{F10 Benchmark Function}
    \label{fig:sub2}
  \end{subfigure}
  %\caption{Two images side by side}
  \label{fig:test}
\end{figure}

\begin{figure}[htbp]
  \centering
  \begin{subfigure}{.5\textwidth}
    \centering
    \includegraphics[width=.9\linewidth]{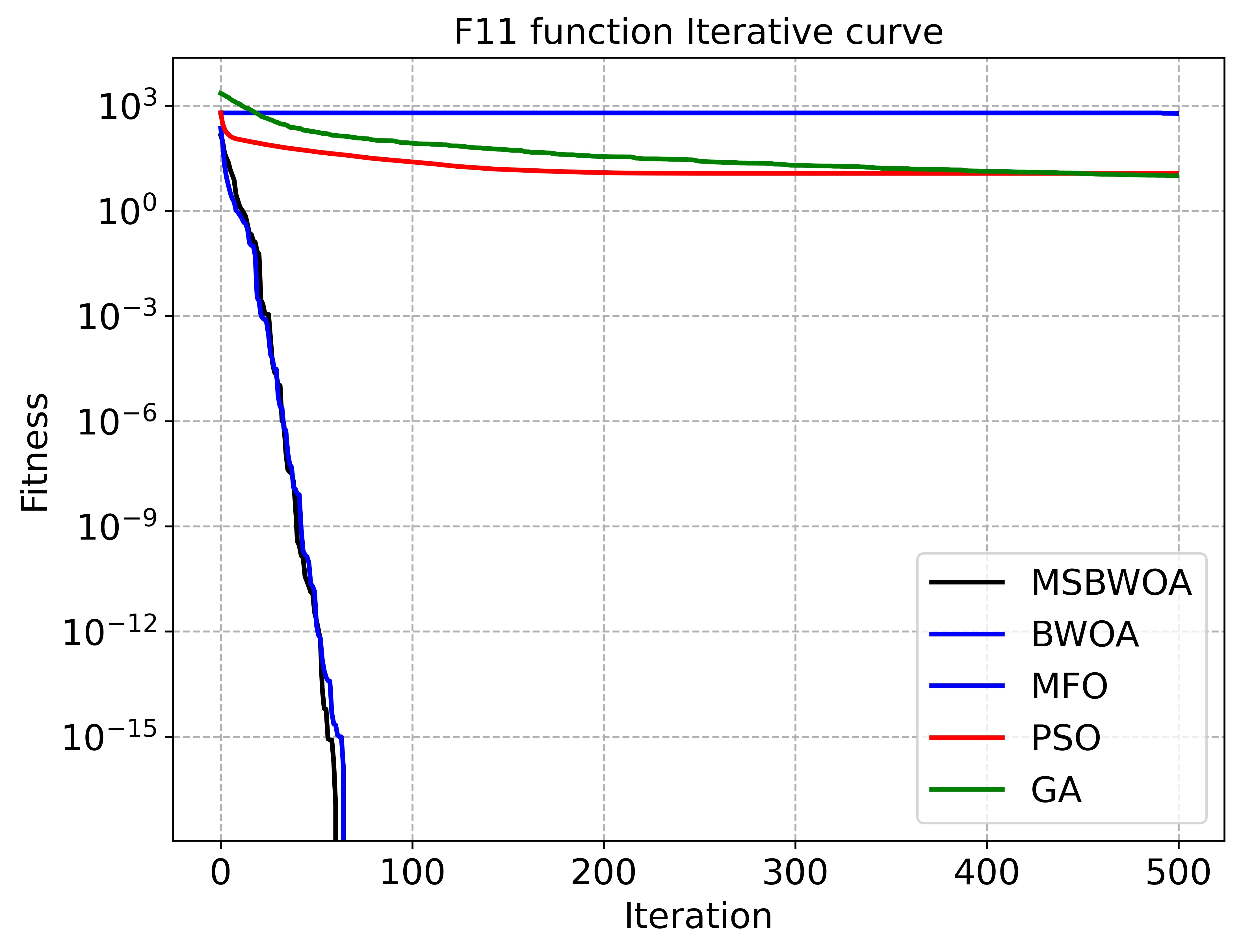}
    \caption{F11 Benchmark Function}
    \label{fig:sub1}
  \end{subfigure}%
  \begin{subfigure}{.5\textwidth}
    \centering
    \includegraphics[width=0.9\linewidth]{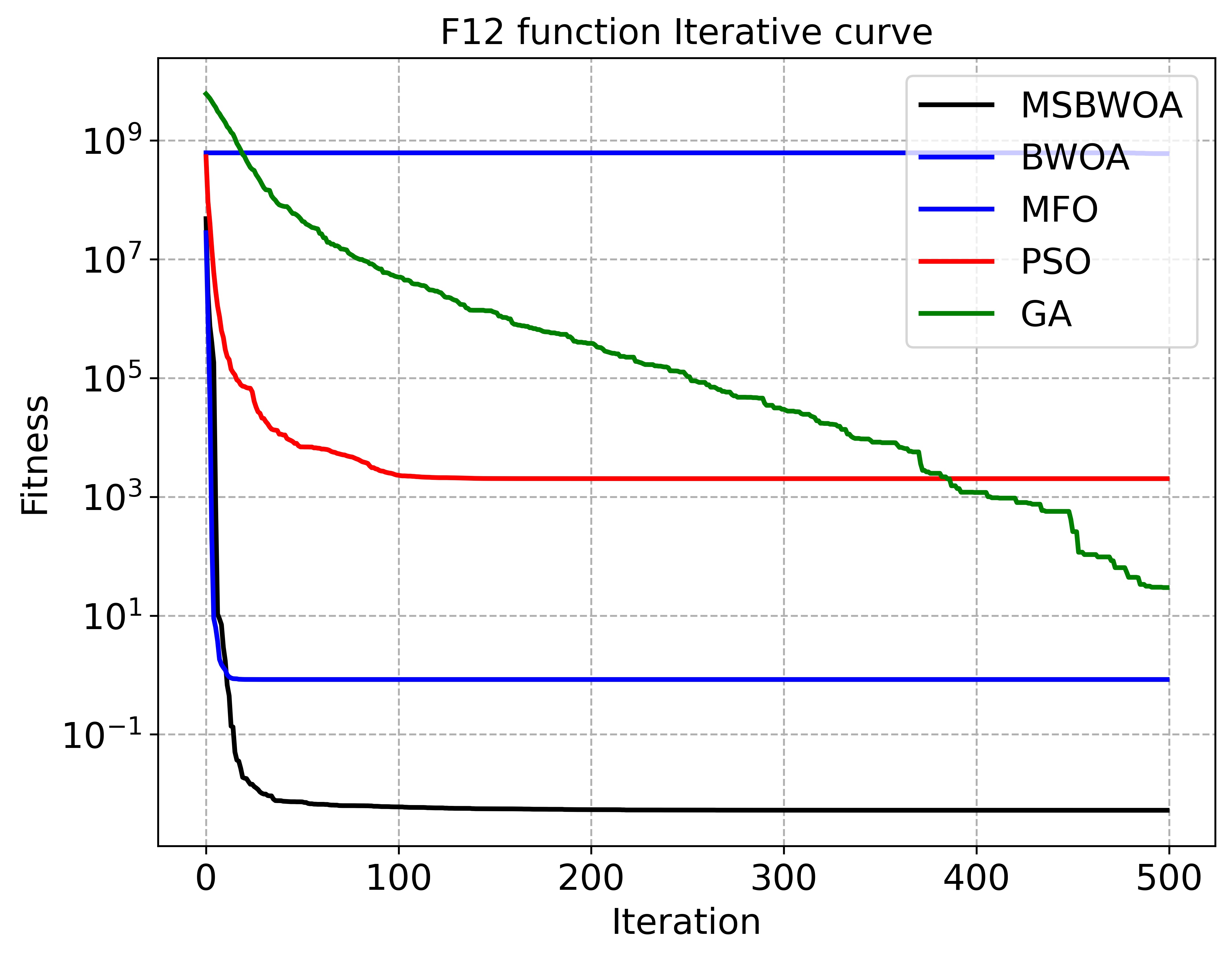}
    \caption{F12 Benchmark Function}
    \label{fig:sub2}
  \end{subfigure}
  %\caption{Two images side by side}
  \label{fig:test}
\end{figure}

\begin{figure}[htbp]
  \centering
  \begin{subfigure}{.5\textwidth}
    \centering
    \includegraphics[width=.9\linewidth]{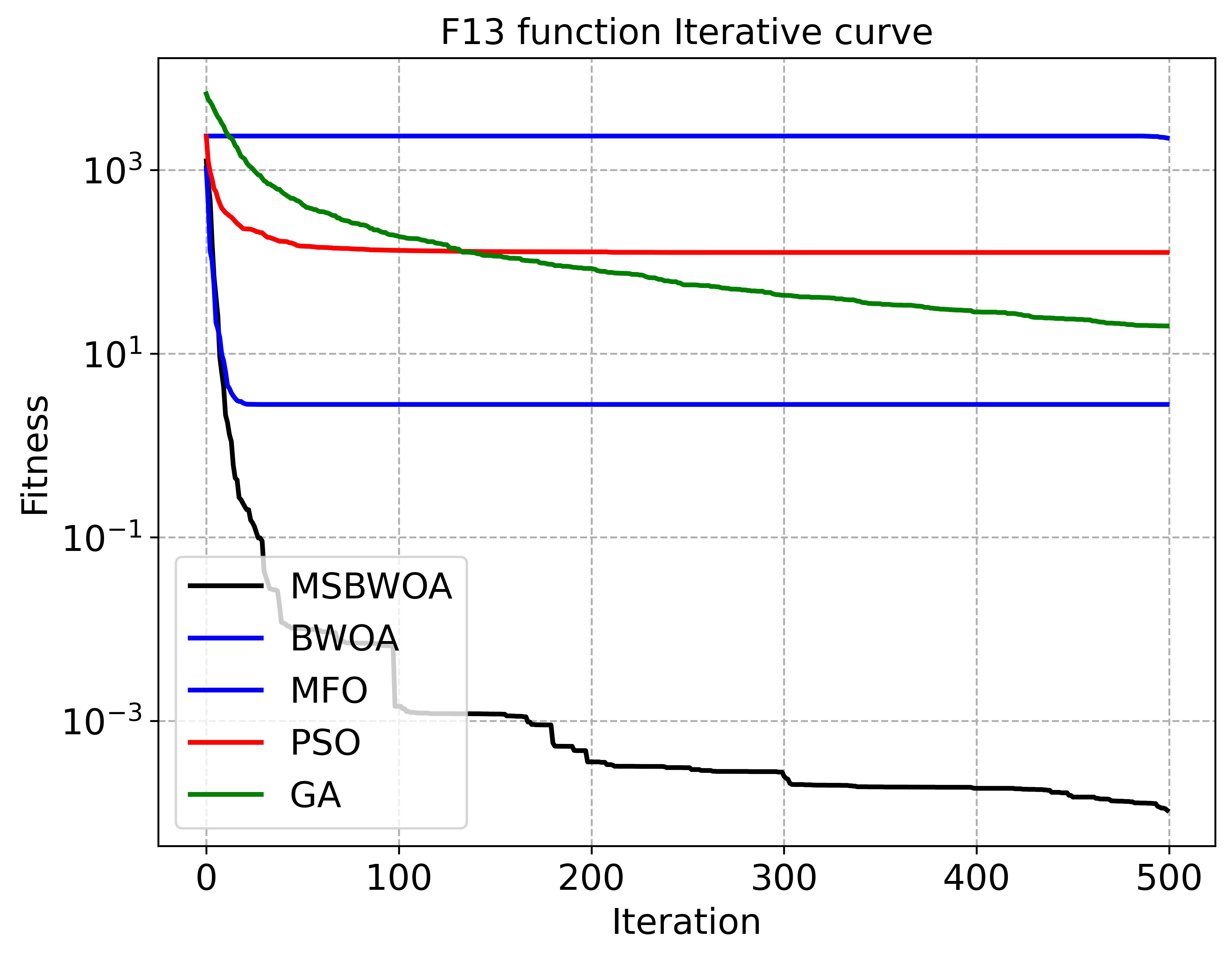}
    \caption{F13 Benchmark Function}
    \label{fig:sub1}
  \end{subfigure}%
  \begin{subfigure}{.5\textwidth}
    \centering
    \includegraphics[width=0.9\linewidth]{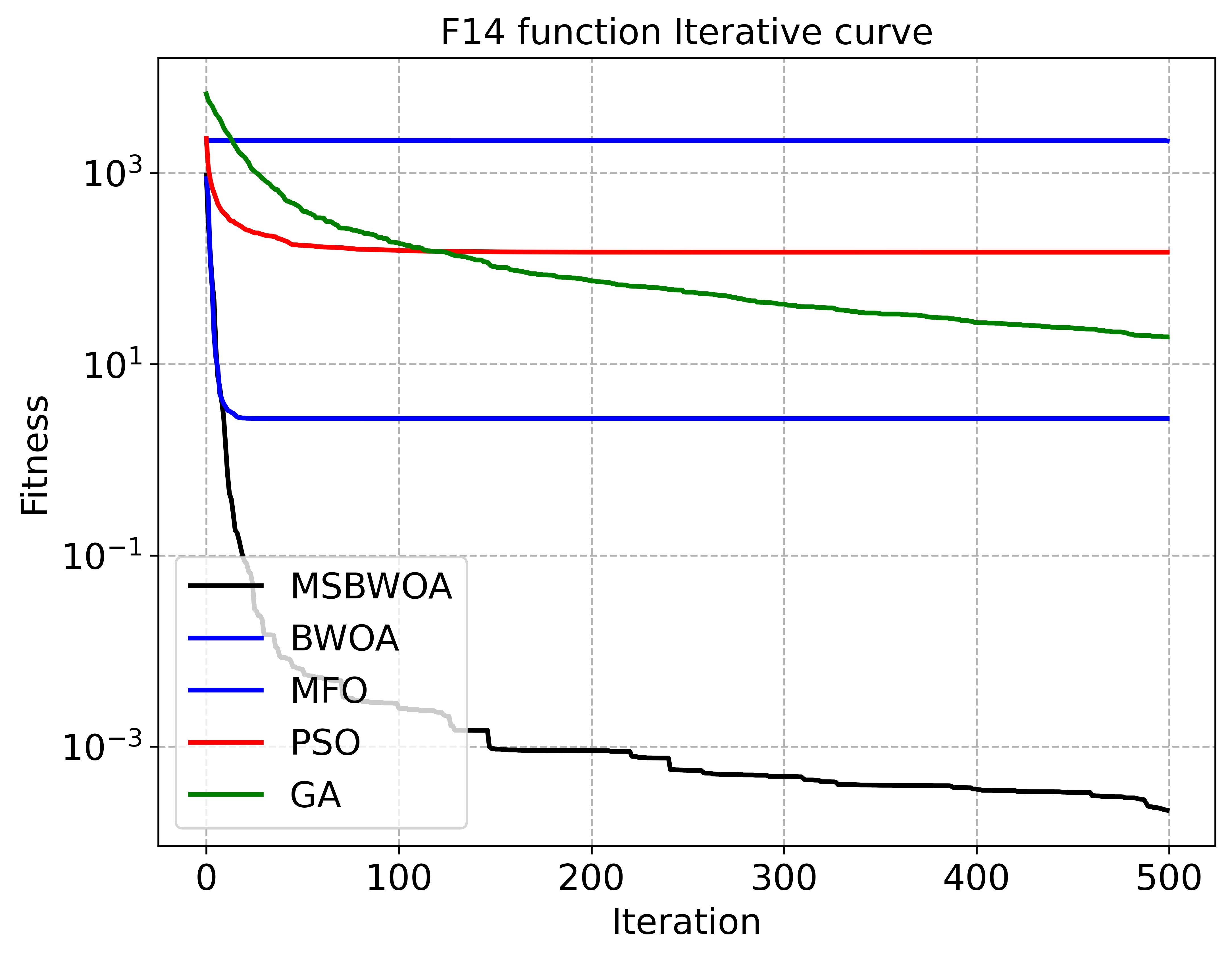}
    \caption{F14 Benchmark Function}
    \label{fig:sub2}
  \end{subfigure}
  %\caption{Two images side by side}
  \label{fig:test}
\end{figure}

\begin{figure}[htbp]
  \centering
  \begin{subfigure}{.5\textwidth}
    \centering
    \includegraphics[width=.9\linewidth]{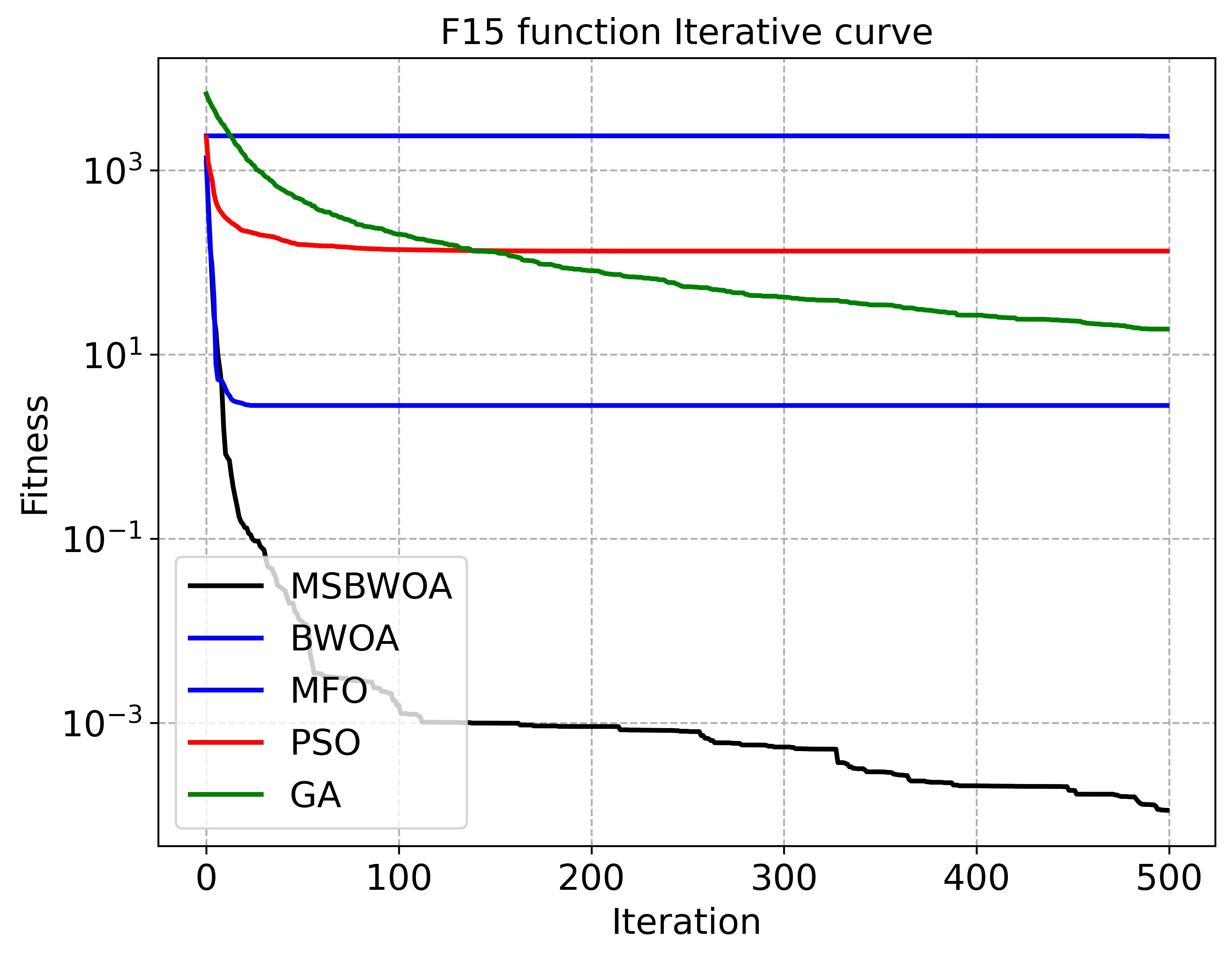}
    \caption{F15 Benchmark Function}
    \label{fig:sub1}
  \end{subfigure}%
  \begin{subfigure}{.5\textwidth}
    \centering
    \includegraphics[width=0.9\linewidth]{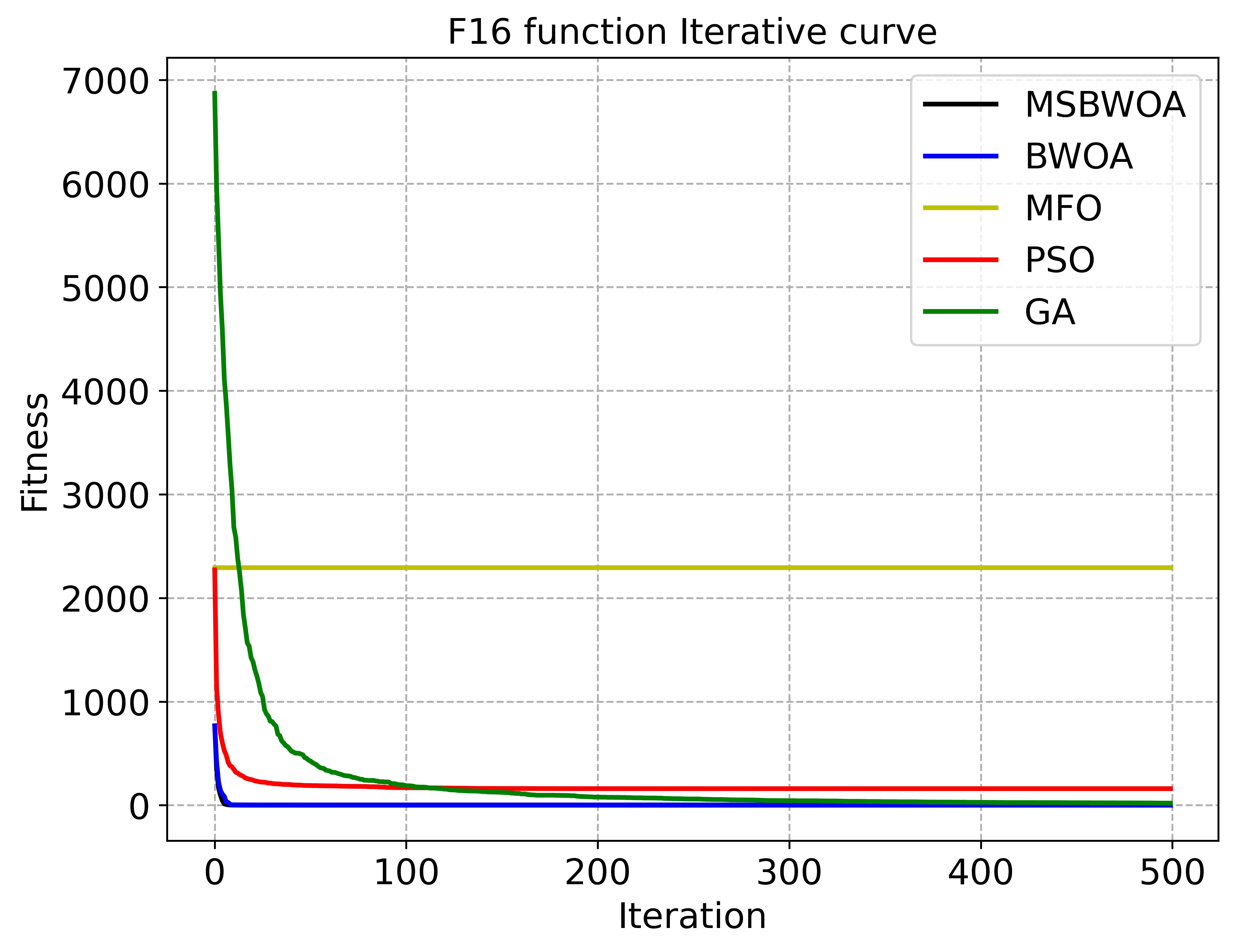}
    \caption{F16 Benchmark Function}
    \label{fig:sub2}
  \end{subfigure}
  %\caption{Two images side by side}
  \label{fig:test}
\end{figure}

\begin{figure}[htbp]
  \centering
  \begin{subfigure}{.5\textwidth}
    \centering
    \includegraphics[width=.9\linewidth]{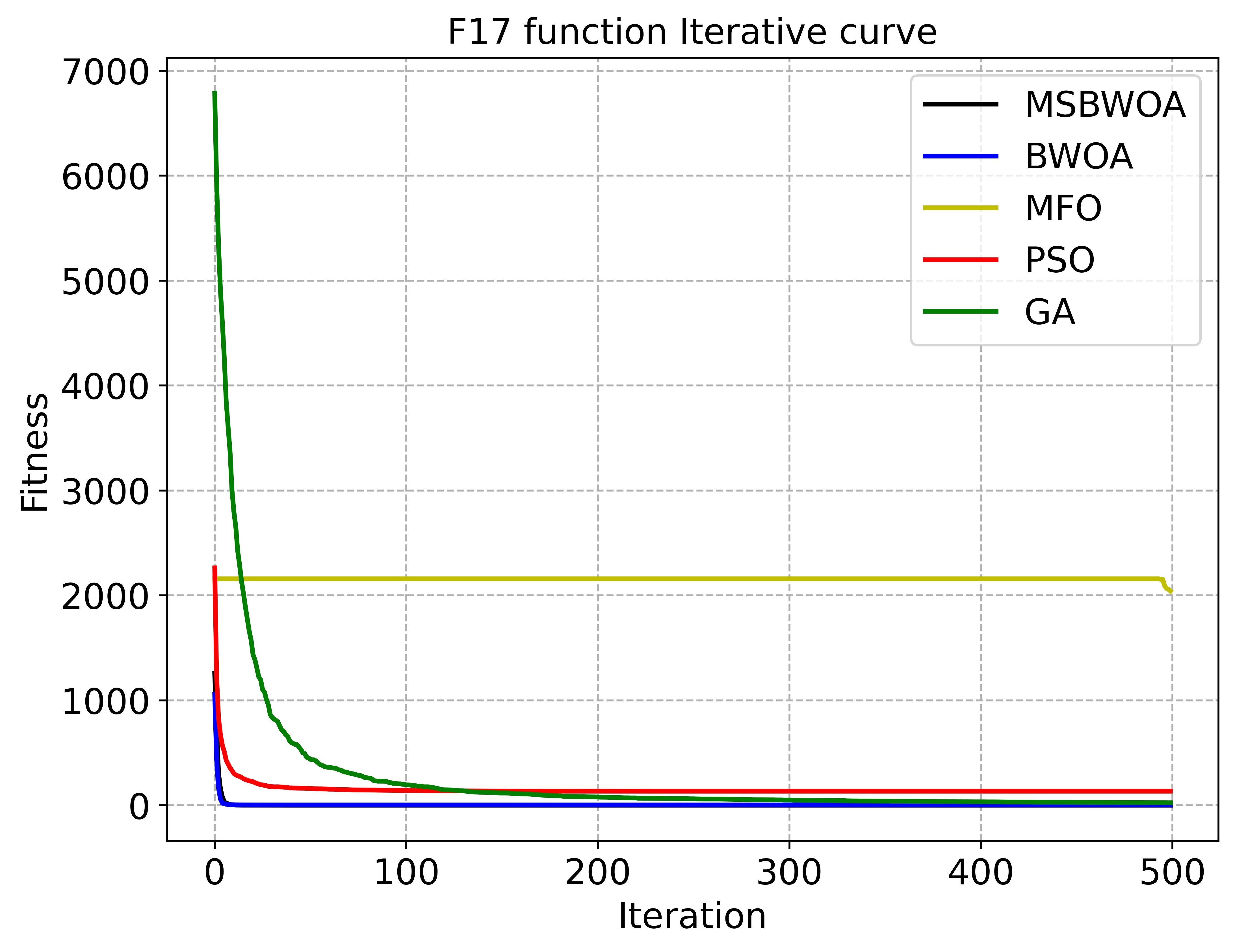}
    \caption{F17 Benchmark Function}
    \label{fig:sub1}
  \end{subfigure}%
  \begin{subfigure}{.5\textwidth}
    \centering
    \includegraphics[width=0.9\linewidth]{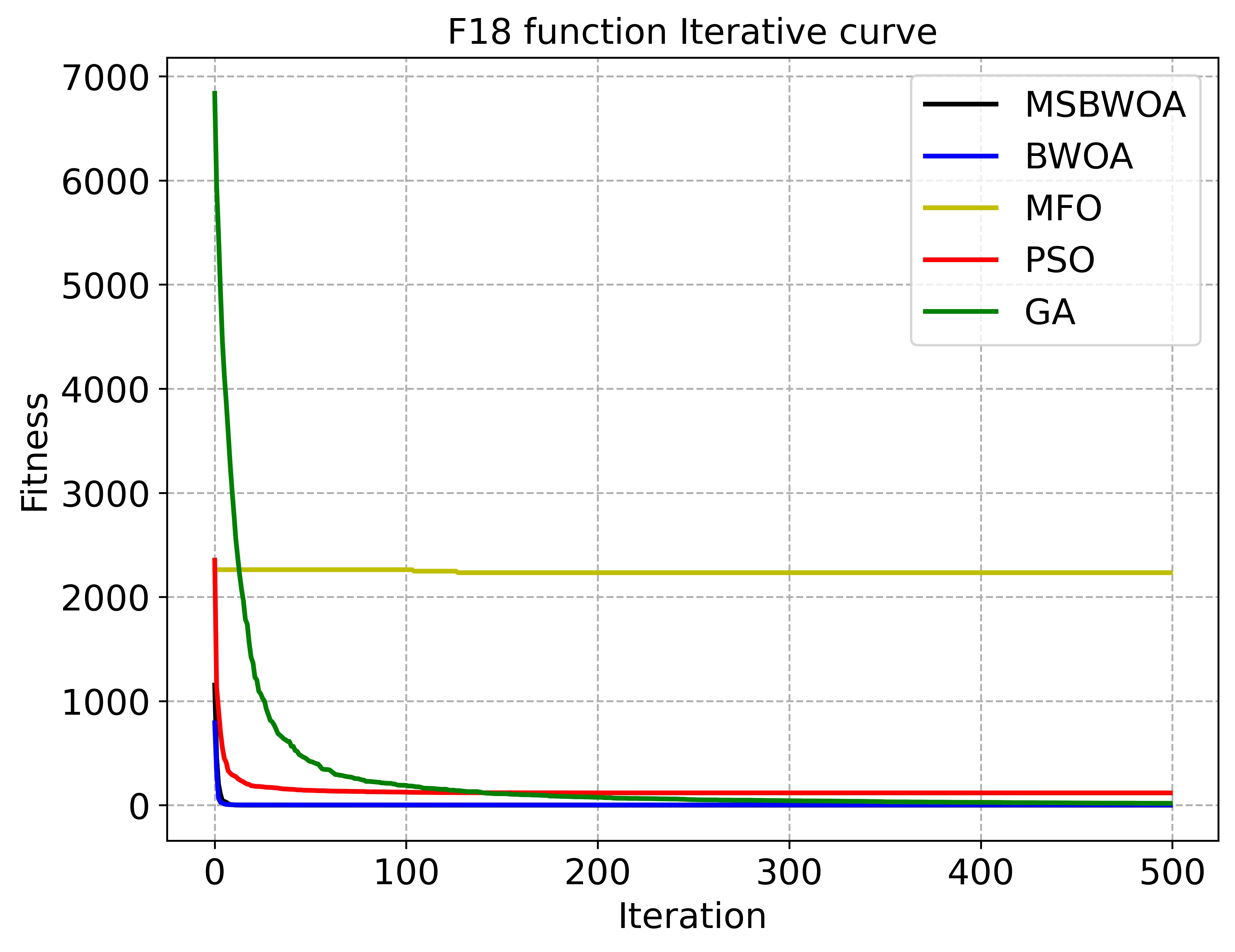}
    \caption{F18 Benchmark Function}
    \label{fig:sub2}
  \end{subfigure}
  %\caption{Two images side by side}
  \label{fig:test}
\end{figure}

\begin{figure}[htbp]
  \centering
  \begin{subfigure}{.5\textwidth}
    \centering
    \includegraphics[width=.9\linewidth]{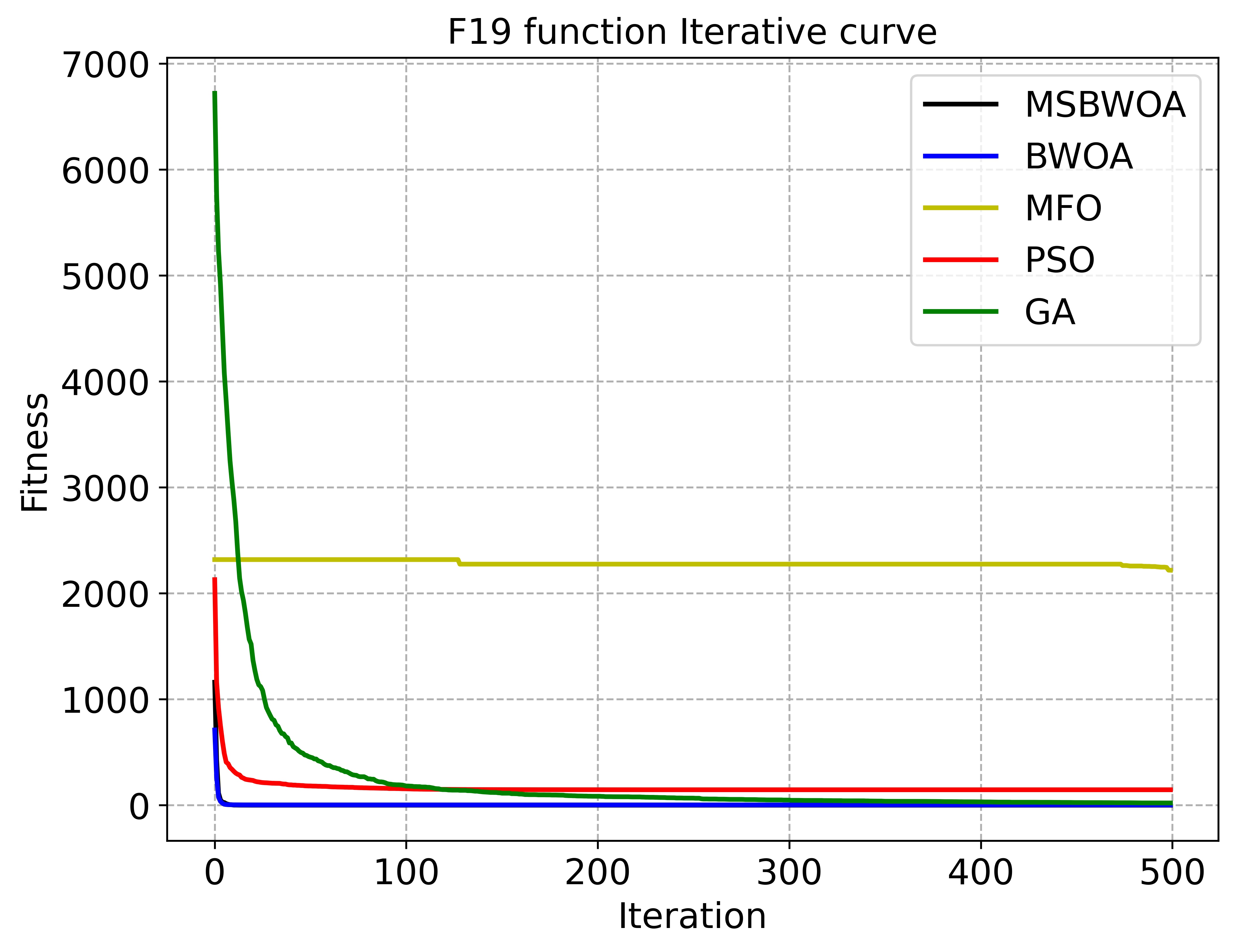}
    \caption{F19 Benchmark Function}
    \label{fig:sub1}
  \end{subfigure}%
  \begin{subfigure}{.5\textwidth}
    \centering
    \includegraphics[width=0.9\linewidth]{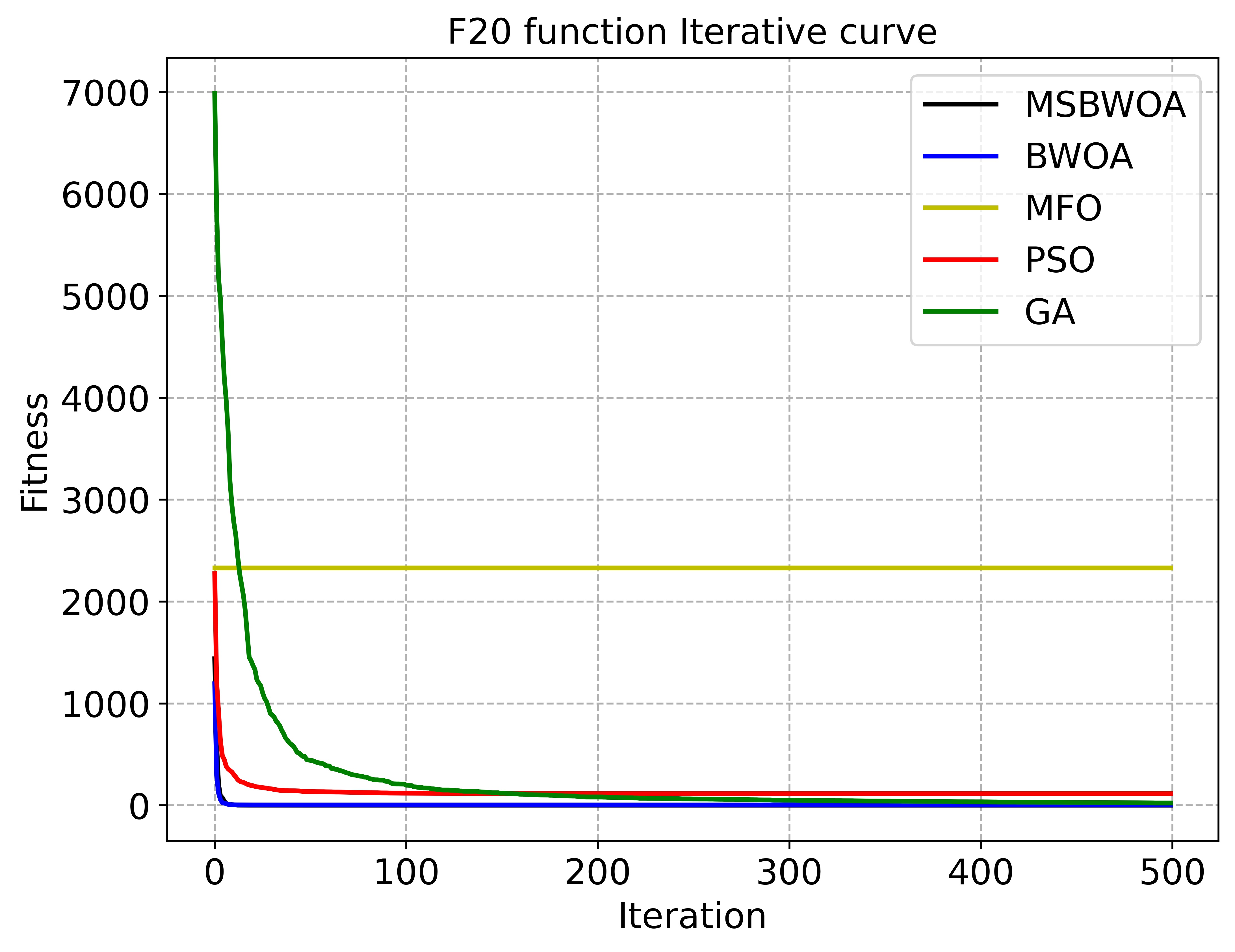}
    \caption{F20 Benchmark Function}
    \label{fig:sub2}
  \end{subfigure}
  %\caption{Two images side by side}
  \label{fig:test}
\end{figure}

\begin{figure}[htbp]
  \centering
  \begin{subfigure}{.5\textwidth}
    \centering
    \includegraphics[width=.9\linewidth]{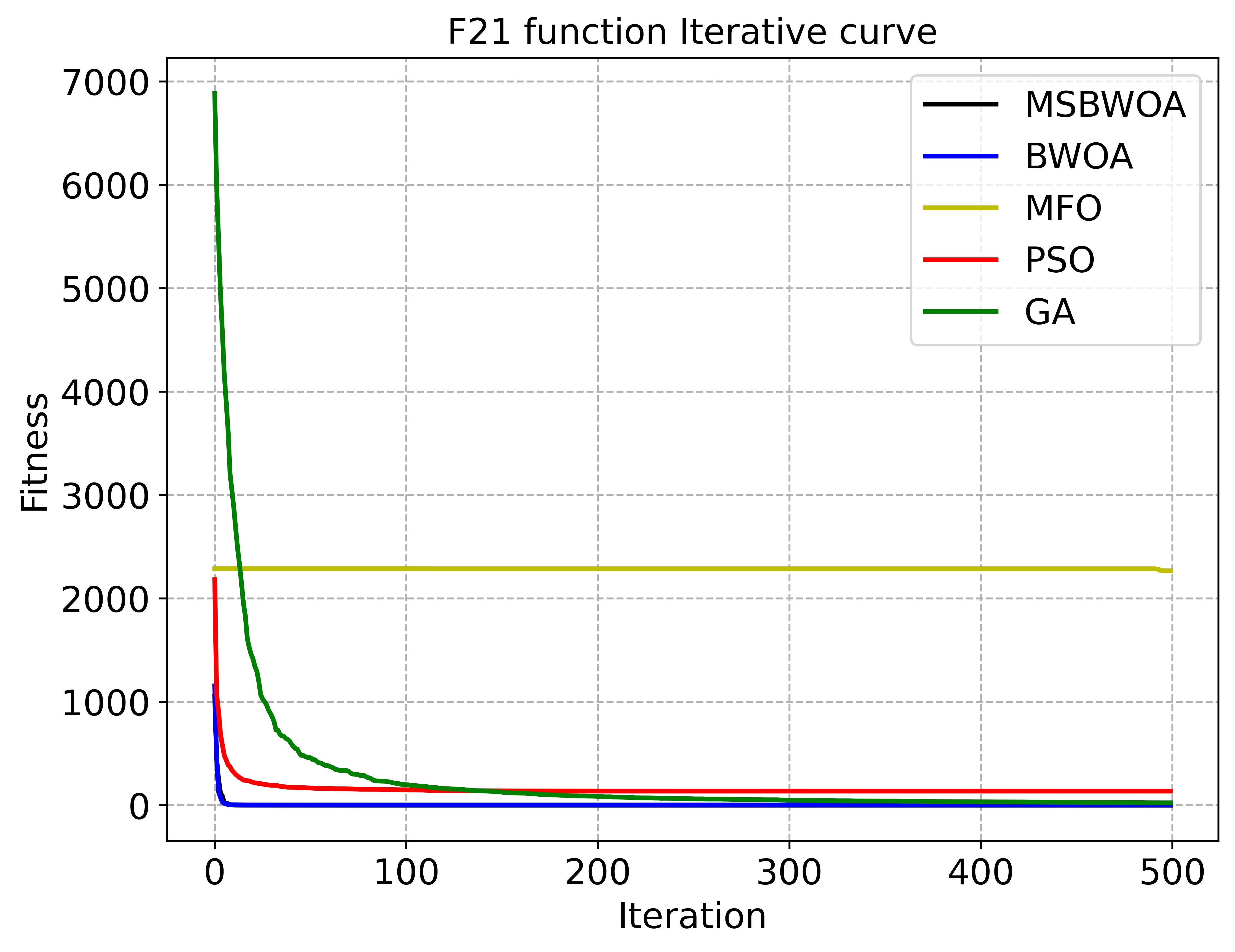}
    \caption{F21 Benchmark Function}
    \label{fig:sub1}
  \end{subfigure}%
  \begin{subfigure}{.5\textwidth}
    \centering
    \includegraphics[width=0.9\linewidth]{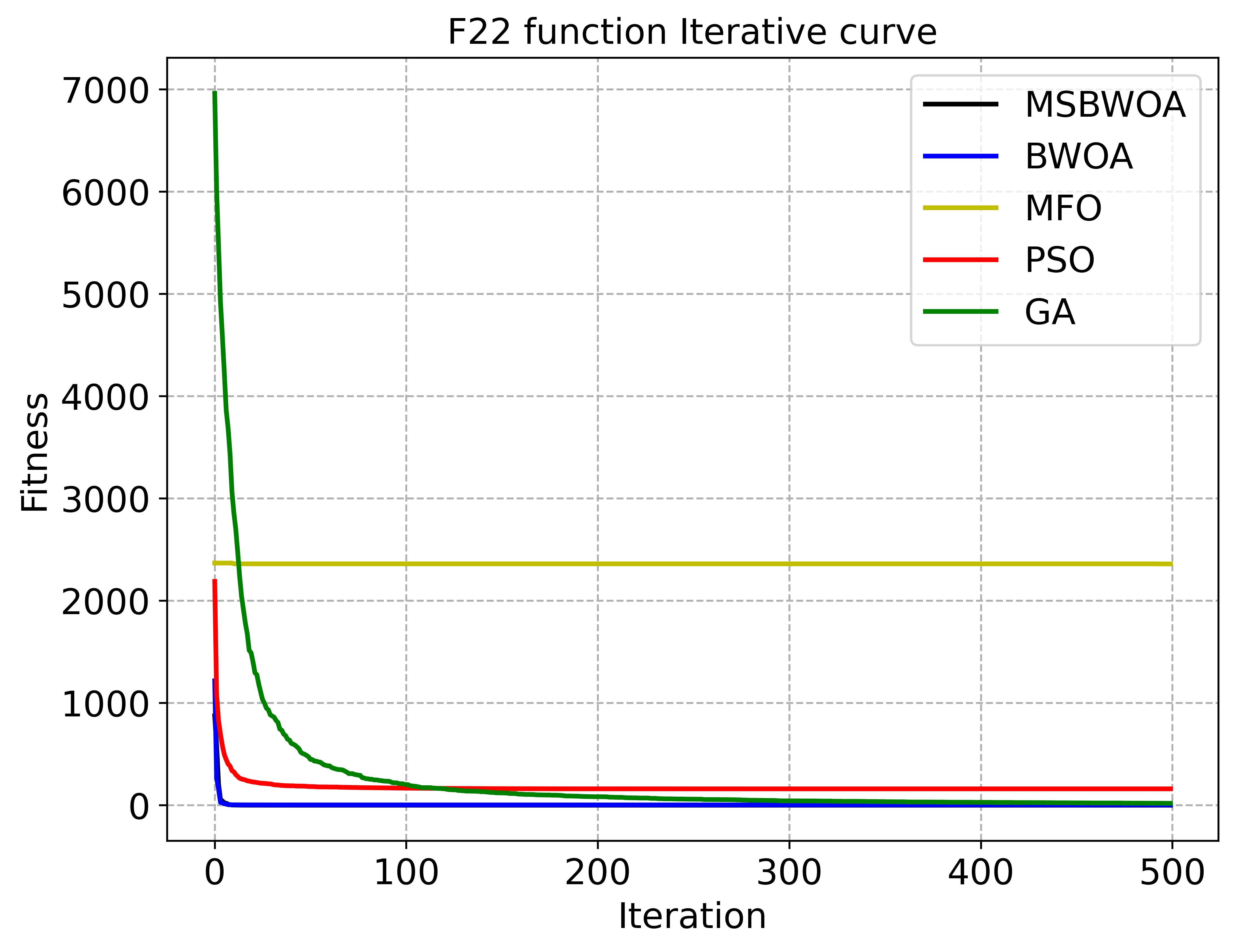}
    \caption{F22 Benchmark Function}
    \label{fig:sub2}
  \end{subfigure}
  %\caption{Two images side by side}
  \label{fig:test}
\end{figure}

\begin{figure}[htbp]
  \centering
  \begin{subfigure}{.5\textwidth}
    \centering
    \includegraphics[width=.9\linewidth]{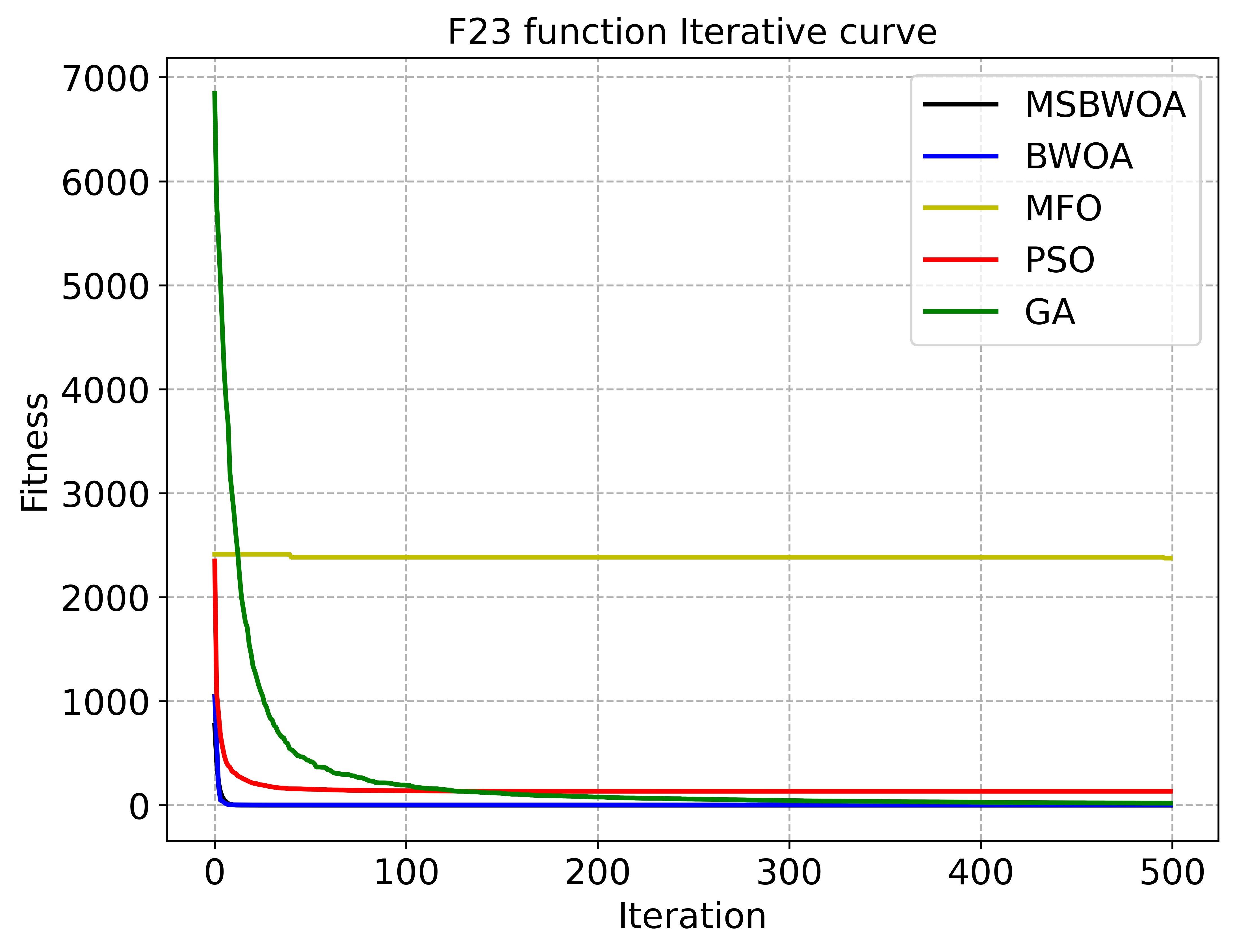}
    \caption{F23 Benchmark Function}
    \label{fig:sub1}
  \end{subfigure}%
  \begin{subfigure}{.5\textwidth}
    \centering
    \includegraphics[width=0.9\linewidth]{Inertial_Weight.jpg}
    \caption{F10 Benchmark Function}
    \label{fig:sub2}
  \end{subfigure}
  %\caption{Two images side by side}
  \label{fig:test}
\end{figure}

\newpage
\bibliographystyle{apacite}
\bibliography{sample}

\end{document}